\documentclass[a4paper,conference]{IEEEtran}
\usepackage{times}
\usepackage{epsfig}
\usepackage{graphicx}
\usepackage{caption}
\usepackage{subcaption}
\usepackage{amsmath}
\usepackage{amssymb}
\usepackage{algorithm}
\usepackage{algorithmic}
\ifCLASSINFOpdf
\else
\fi
\begin{document}
%
\title{Aggregated Sparse Attention for Steering Angle Prediction}

\author{\IEEEauthorblockN{
		Sen He,
		Dmitry Kangin, 
		Yang Mi
		and Nicolas Pugeault}
\IEEEauthorblockA{Department of Computer Sciences,
University of Exeter, 
Exeter, EX4 4QF\\
Email: \{sh752, D.Kangin, ym310, N.Pugeault\}@exeter.ac.uk}}


%


\maketitle

\begin{abstract}
In this paper, we apply the attention mechanism to autonomous driving for steering angle prediction. We propose the first model, applying the recently introduced sparse attention mechanism to visual domain, as well as the aggregated extension for this model. We show the improvement of the proposed method, comparing to no attention as well as to different types of attention.
\end{abstract}


%
\IEEEpeerreviewmaketitle

\section{Introduction}\label{sec:intro}
Consider a human driving a car on a countryside road. The driver's brain is subjected to a continuous flow of large quantities of visual information, interpreting it in real time to provide fast, precise and reliable control of the vehicle. 
An essential mechanism that allows such an efficient and fast processing of information is \textit{visual attention}, which has been extensively studied by psychologists. 
%
Early computational models of attention, inspired by the seminal work of Itti \& Koch \cite{IttiKoch2000}, focused on the top-down mechanism that elicit eye movements when subjects perform a visual search of objects on images. 
The aim of such models is to estimate from an image a so-called \textit{saliency map}: an estimate of how likely are the subject's eyes to look at image locations given the patterns it contains (see, eg, \cite{kruthiventi2017deepfix,pan2016shallow}). Saliency models can either be engineered based on properties of images, or learnt from eye tracking records of human subjects. In both case, the quality of saliency models is estimated by comparing their prediction with actual eye fixations on dataset of images. 
Although such approaches can predict fairly well the eye fixations of human subjects when asked to perform a visual search task, their predictiveness is much worse when the subjects are performing an \textit{active} task, such as playing video games or driving~\cite{BorjiEtAl2014,pugeault2015much}. 
More recently, several groups have proposed to learn attention not by mimicking the gaze of human subjects, but by optimising a system's performance at a specific task \cite{limam,sharma2015action}. In contrast to saliency, that is purely bottom-up, such models are explicitly task dependent. 

This article proposes a novel attention mechanism for convolutional neural networks that is based on learning a task-specific sparse attention mechanism. In particular, we focus on the challenging task of predicting steering angle from visual input only \cite{pugeault2015much}.  We demonstrate that such a sparse attention focusing leads to better performance. Moreover, we provide experimental evidence that such an attention model is very sensitive to initial conditions and demonstrate that an ensemble of sparse attentional models can significantly improve not only the robustness of the learning process, but also overall performance. 

The rest of this article is organised as follows: 
Section~\ref{sec:related} reviews the use of attention mechanism in computer vision as well as steering angle prediction; 
Section~\ref{sec:methods} provides the detailed methodology used in this work; 
Experimental results, together with comparison are presented in Section~\ref{sec:results}; 
and conclusions are drawn in Section~\ref{sec:conclusion}.

\begin{figure}
	\centering
	\includegraphics[width=0.8\columnwidth]{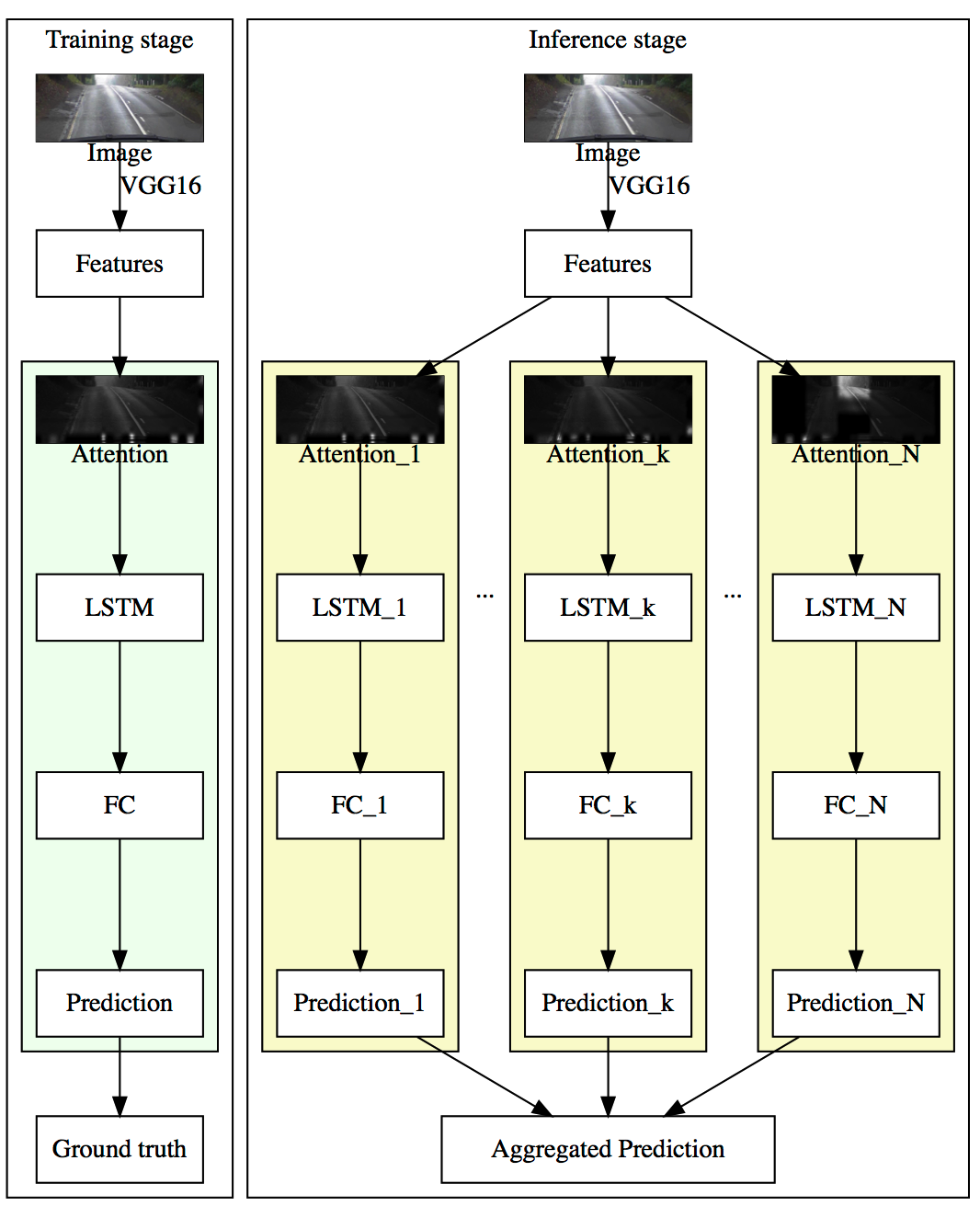}
	\caption{Architecture of the proposed aggregated attention model. }
	\label{fig:train}
\end{figure}

\section{Related Work}\label{sec:related}
A broad range of attention models have been proposed over the years in the literature. This article is concerned in particular with the problem of \textit{task-dependent attention}, where the focusing of attention is optimised to \textit{improve a system's performance at a given task}. 
This is in contrast to saliency models which are designed to mimic human subjects' gaze patterns irrespective of the tasks demands. 
Existing models of task-dependent attention for neural networks can be classified in two groups: soft attention and hard attention. 

In \textit{soft attention}, the visual input is processed by a pre-trained convolutional neural network, and the output of the top convolutional layer is encoded by a feature tensor. 
Soft attention consists in weighing the feature tensor with an attention matrix that encodes the relative importance of all locations in the feature tensor. This weighted tensor is fed to another network (ie, a fully connected or a recurrent neural network) to optimise the desired task. The attention matrix is therefore learnt from the task and normalised using a softmax function. 
Li et al~\cite{limam} used soft attention to develop a multilevel attention model for video captioning. Their model uses two attention layers. The first layer models \textit{region-level attention}, which encodes the importance of each region in a frame. The second attention layer models \textit{frame-level attention}, which encodes the importance of each frame in a short video.
Sharma et al~\cite{sharma2015action} proposed a soft attention model for action recognition. In their model, the output of a pre-trained deep convolutional neural network is fed to a \textit{long short-term memory (LSTM)} network to output the action as well as the attention matrix.
Xu et al~\cite{xu2015show} used soft attention for image captioning. Their model also uses a LSTM network that generates an attention matrix at each time step and generating sentences to describe the input image. 

One limitation of soft attention is that it only reweighs the convolutional features and therefore everything is always attended to, although not with the same relative importance (hence \textit{soft} attention). 
In contrast, \textit{hard attention} models only process part of the input, which is assumed to be the most important region. 
Because hard attention is not differentiable, it is more challenging to optimise. 
Mnih et al \cite{mnih2014recurrent} proposes to learn hard attention from reinforcement learning. There are two crucial component in their network: The first one is a glimpse sensor, which can be used to extract a retina-like representation centred at a given location in the input; The second component is a glimpse network, it is used to process the retina-like representation extracted from the glimpse sensor, and the processed information is then fed into a recurrent neural network (RNN) which estimates the attention focus for the glimpse sensor at the next iteration. 

This article is especially concerned with active tasks, and in particular the problem of estimating steering from vision. 
Pugeault \& Bowden~\cite{pugeault2015much} developed a pre-attentive model using gist~\cite{oliva2001modeling} and random forests, while the deep network models are CNN or CNN$+$ LSTM based, Bojarski et al~\cite{bojarski2016end} used a convolutional neural network to map images to steering command.
Du et al~\cite{duself} explore two different models for steering angle prediction. The first is a 3D convolutional model with residual connections and LSTM cell. The second is uses transfer learning to fine-tune a pre-trained CNN and predict steering angles for individual images. 

Much less work has been done on the application of attention mechanism in autonomous driving. 
In this article: i) we propose a new sparse attention model, based on the sparsemax function \cite{martins2016softmax}, yields better performance; ii) we demonstrate that bagging multiple sparse attention models can provide a significant performance improvement over single models; iii) we show that the proposed architecture performs better than the state-of-the-art soft attention model, CNN, CNN+LSTM for steering angle prediction. 

\section{Regression of steering angles with attention}\label{sec:methods}
This section describes the proposed sparse attention model: First, we present the overall architecture in Figure~\ref{fig:train}; we then describe the LSTM model and sparse attention formulation for steering regression; and finally, the proposed model aggregation approach.  

\subsection{Feature Extraction}
The deep convolutional neural networks have achieved great success in computer vision due to its ability to learn hierarchical features. In our model, we extract the feature for each frame in a driving video using the convolutional part of VGG16~\cite{simonyan2014very}, which was trained for image recognition. After feature extraction, each frame was represented by a tensor of shape $M\times N\times K$ determined by the input size. We refer to feature tensor as a feature cube with $M\times N$ locations, and each location was represented by a feature vector of $K$ elements:
\begin{align}
X = \left[ X_{1},X_{2},\cdot \cdot \cdot,X_{M\times N}\right]
\end{align}
The feature extraction part is fixed during the experiment and not fine-tuned.
\subsection{LSTM}
In order to take into account the previous context to predict steering angle, the recurrent neural network was used. Recurrent neural network (RNN) can process the time sequence by remembering the needed information and forgetting the redundant. Long Short Term Memory (LSTM) networks \cite{gers1999learning} are a kind of gated RNN, which can avoid the gradient vanishing or exploding problems encountered by standard RNNs.

\subsection{Sparse Attention}
A fundamental limitation of soft attention is that all image regions are in effect attended to at all times: Their importance is merely reweighed by the attention model. This is contrary to the very intent of attention learning.  

In this work, we propose to mitigate this limitation by implementing a \textit{sparse attention mechanism} based on~\cite{martins2016softmax}, but extending it to visual inputs. 
The output of the attention transformation is defined as
\begin{align}
X_{weighted} = X \cdot A,
\end{align}
where the elements of sparse attention matrix $A$ sums to 1, and it is determined by the feature cube of the current input frame and the model hidden state in the last time step and normalised by the sparsemax function:
\begin{align}
A = \text{sparsemax}(\text{tanh}(W_{f}X + W_{h}H + b))
\end{align}
where the $W_{f}$ is weight for the current input frame's feature, $W_h$ is the weight for the model hidden state, $H$ is the hidden state of the model, both of the weights are learned during training to form the attention matrix, the sparsemax function which is defined by \cite{martins2016softmax} in Algorithm \ref{Sparsemax_algorithm}.

\begin{algorithm}
\caption{sparsemax}
\label{Sparsemax_algorithm}
$\boldsymbol{Input: z}$\\
Sort $z_{(1)} \geq \cdots \geq z_{(M\times N)}$\\
Find $k(z)$ $:=$ max  \{ $k \in \left[ M\times N\right] | 1+kz_{(k)} > \sum_{j \leq k}{z_{(j)}}$ \}\\
Define $\tau (z)$ $=$ $\frac{(\sum_{j \leq k_{(z)}}z_{(j)})-1}{k_{(z)}}$\\
$\boldsymbol{Output: p}$ s.t. $p_{i}$ $=$ $\left[z_{i} - \tau(z)\right]_{+}$
\end{algorithm}

One can see that the sparsemax function is not continuous. More importantly, compared to the softmax function, it has the ability to inhibit the unimportant but enhance the significant elements of the input~\cite{martins2016softmax}.
The final prediction is generated by a two layer fully connected layers (FCN):
\begin{align}
S(t) = W_{fcn2} (W_{fcn1}H_{t} + b_{fcn1}) + b_{fcn2}
\end{align}
where, $S(t)$ is the predicted steering angle, $W_{fcn1}$ and $W_{fcn2}$ are the weights of each fully connected layer, $b_{fcn1}$ and $b_{fcn2}$ are the bias of each layer, and $H_{t}$ is the output of LSTM.

\subsection{Model Aggregation} \label{Model Aggregation}
Due to the non-continuity of the sparsemax function, we suggest that the result of training a sparse attention model is highly dependent on (random) initialisation. This means that the resulting attention models after training, although converging to similar performance levels, correspond to very different local minima depending on the random initialisation.  
In other words, the same task can afford multiple attention models of similar quality. If those models all capture different aspects of the task, a combination of those models could lead to better performance. 
Therefore, we propose to train a collection of $N$ (we choose $N=3$ in the experiment) randomised attention models. Because model variance can be ensured from the random initialisation, we can train them all using the same dataset (experiments confirmed that training each model on a separate bootstrap samples did not alter the results significantly). At inference time we propose to combine these attention models and average their predictions, similarly to model bagging.

\section{Experimental Results}\label{sec:results}
The advantages of the proposed method are shown using DIPLECS dataset \cite{pugeault2015much}, containing indoor and outdoor scenarios, and \textit{Comma.ai} dataset \cite{santana2016learning}. The proposed method is compared to soft attention and aggregated soft attention (ASA).  Also the method is compared to the gist-based approach \cite{pugeault2015much}, method with no attention and no LSTM (CNN), as well as LSTM without attention (CNN+LSTM). To measure the prediction quality, we use the mean absolute error.

\subsection{Dataset description}
\begin{figure*}[h!]
	\centering
\begin{tabular}{ccc}
	\includegraphics[width=.3\textwidth]{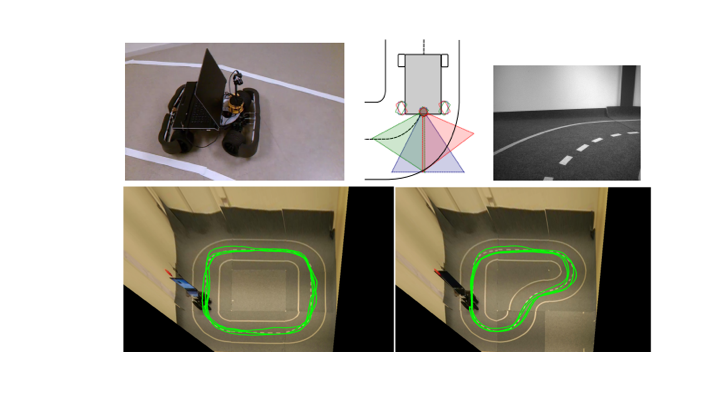}
	&
	\includegraphics[width=.3\textwidth]{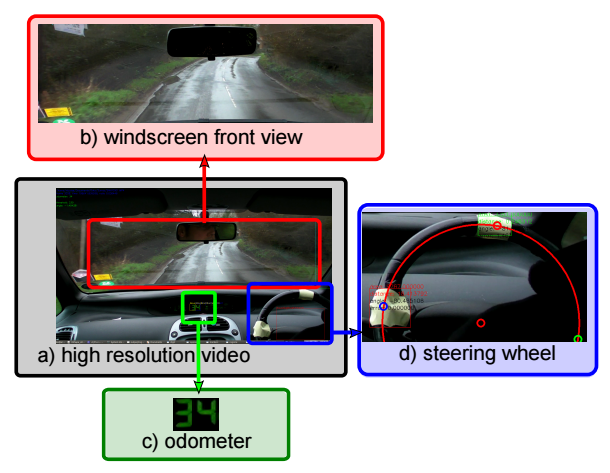}
	&
	\includegraphics[width=.25\textwidth]{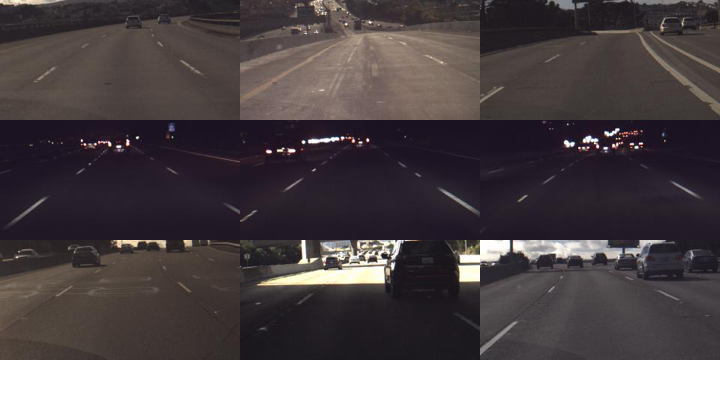}
\end{tabular}
\caption{
	The datasets used in this article: 
	left, the \textit{DIPLECS} indoor dataset (figure reproduced from \cite{pugeault2015much}); 
	middle, the \textit{DIPLECS} indoor dataset (figure reproduced from \cite{pugeault2015much};
	right, the \textit{Comma.ai} dataset \cite{santana2016learning}.
}
\label{fig:datasets}

\end{figure*}

The indoor part of the \textit{DIPLECS} dataset \cite{pugeault2015much} is collected using a radio controlled car (see Figure~\ref{fig:datasets}, left). There are two tracks, P-shaped and O-shaped, eight recordings for each of them from different starting point. For each of the tracks, three recordings are used for training, one for validation, and the rest for testing. 

The outdoor part of the \textit{DIPLECS} dataset \cite{pugeault2015much} contains real world driving scenarios (see Figure~\ref{fig:datasets}, middle), totalling about $47$ minutes of driving, or $84,690$ frames. This dataset has been divided into eight  subsequences of the same length, with junctions removed as causing ambiguity which cannot be resolved using vision-based information. Six subsequences were used to train the model, another two were used for validation and testing. In order to factor out focusing attention on the steering wheel and the mirror, these regions were cropped. 


\textit{Comma.ai} dataset consists of 10 day- and night-time highway driving video clips of variable size, in total $7.5$ hours (see Figure~\ref{fig:datasets}, right). We extract 8 sequences from the dataset, each of them contains $4000$ frames, and use 6 of them to train and the rest for validating and testing. 

\subsection{Training parameters}
Our models are trained using Tensorflow \cite{abadi2016tensorflow} with the $L^1$ norm loss function, we set the learning rate as $10^{-4}$ and use Adam~\cite{kingma2014adam} optimisation method to train the model, all weights to be trained in the model are initialised using Xavier~\cite{glorot2010understanding} initialisation method.

\subsection{Indoor Dataset Results}
\begin{figure*}[h!]
	\begin{center}
		\begin{subfigure}{0.19\textwidth}
			\includegraphics[width=\textwidth]{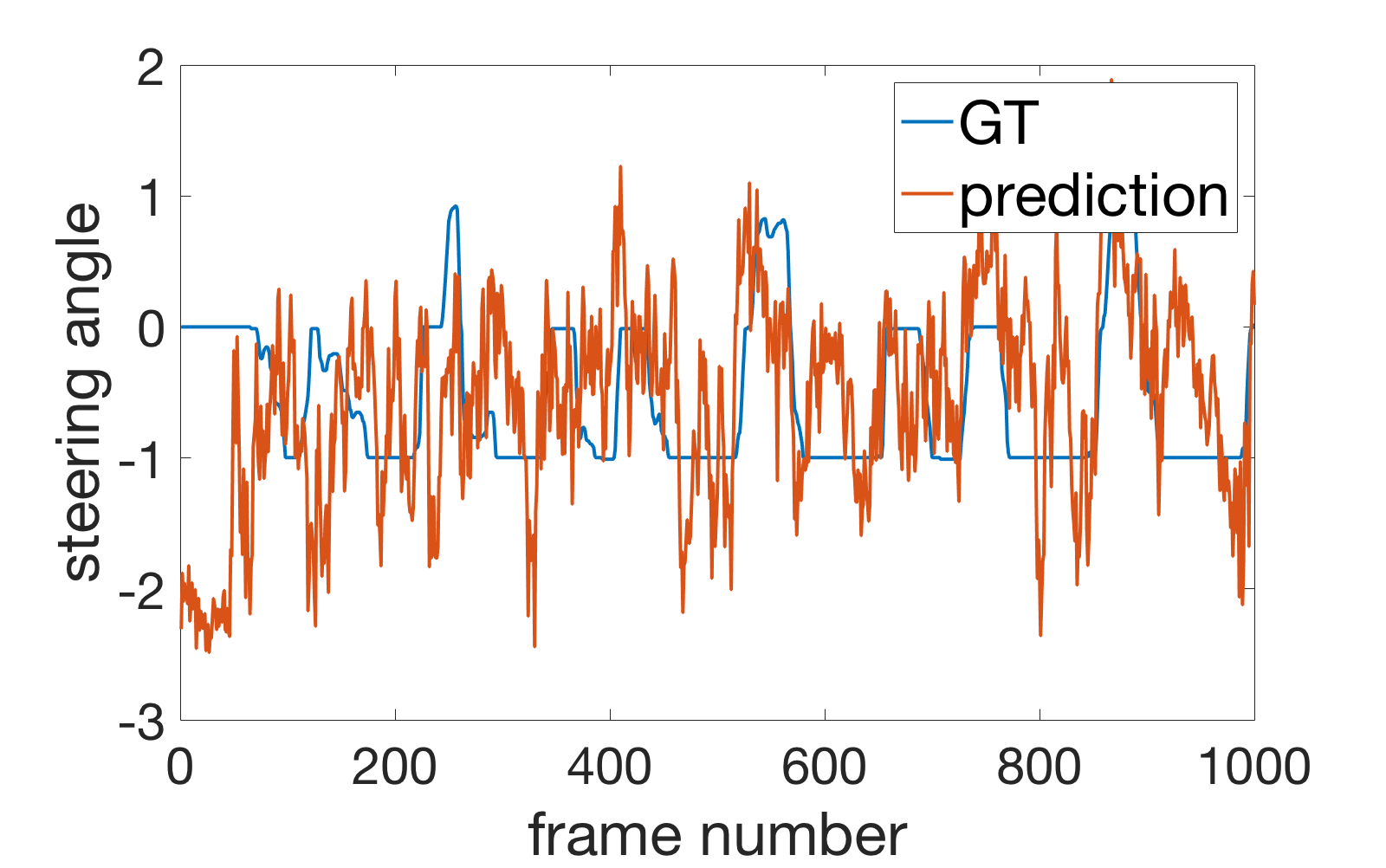}
			\caption{CNN}
		\end{subfigure}
		\begin{subfigure}{0.19\textwidth}
			\includegraphics[width=\textwidth]{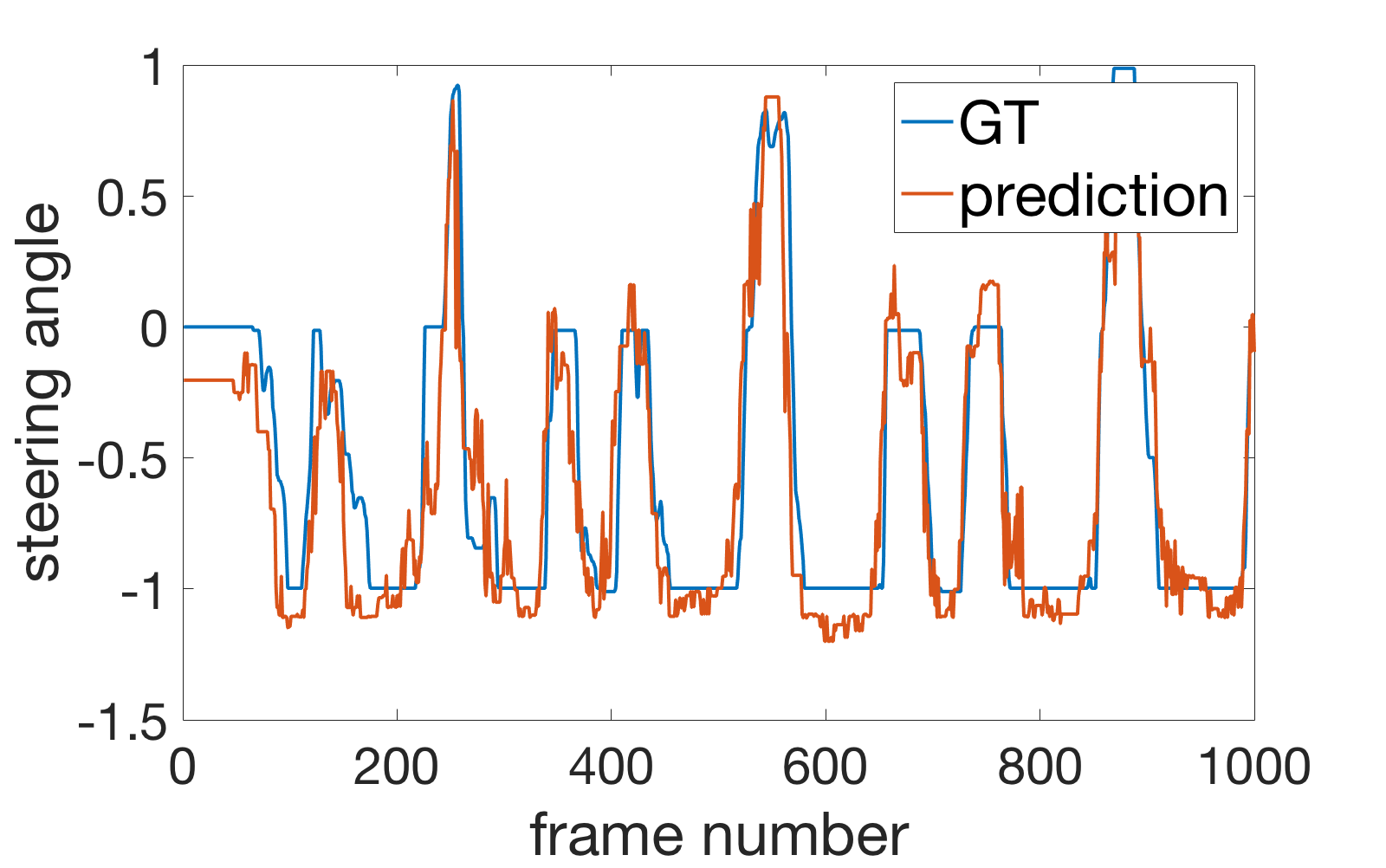}
			\caption{CNN+LSTM}
		\end{subfigure}
		\begin{subfigure}{0.19\textwidth}
			\includegraphics[width=\textwidth]{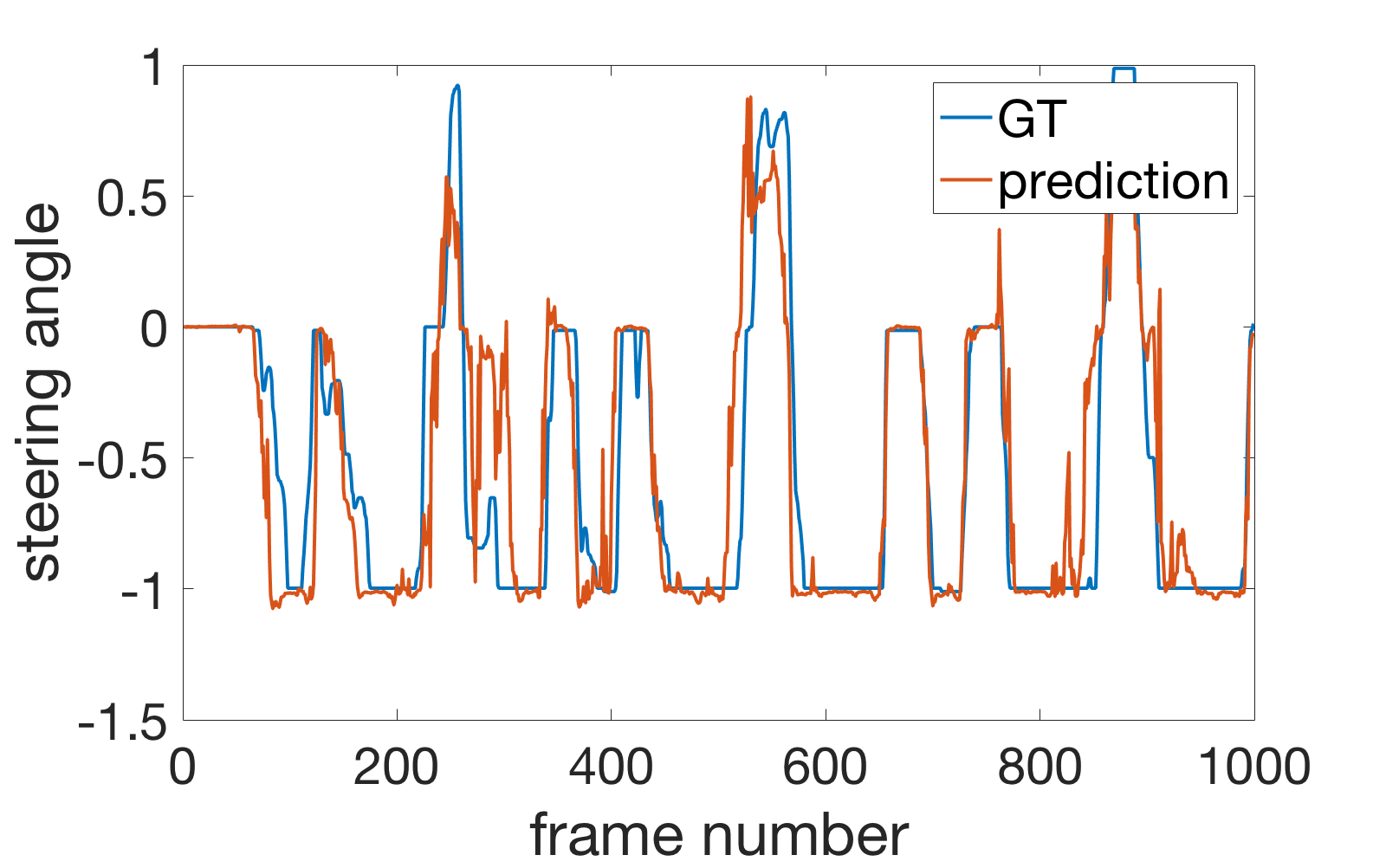}
			\caption{Soft attention}
		\end{subfigure}
		\begin{subfigure}{0.19\textwidth}
			\includegraphics[width=\textwidth]{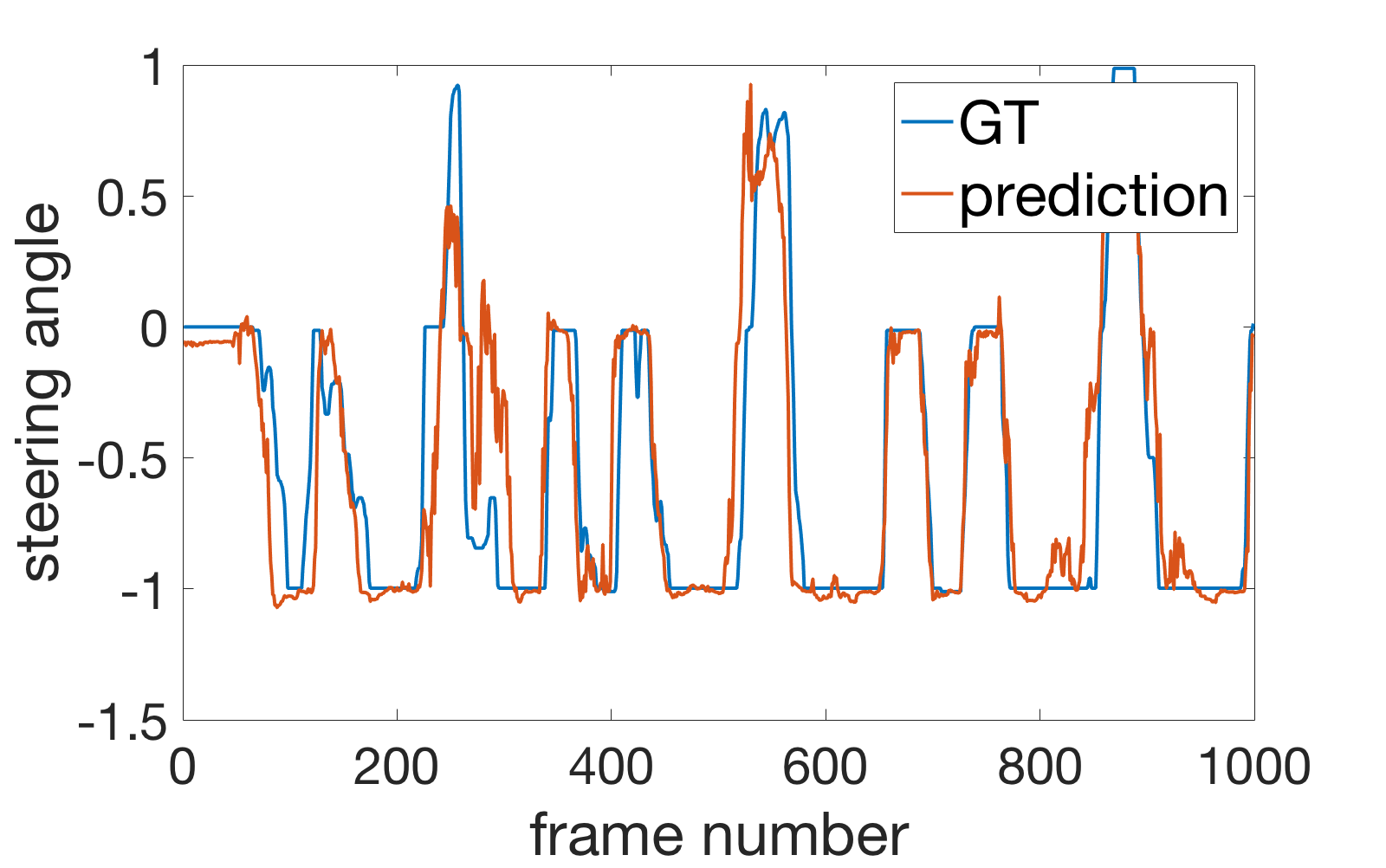}
			\caption{ASA}
		\end{subfigure}
		\begin{subfigure}{0.19\textwidth}
			\includegraphics[width=\textwidth]{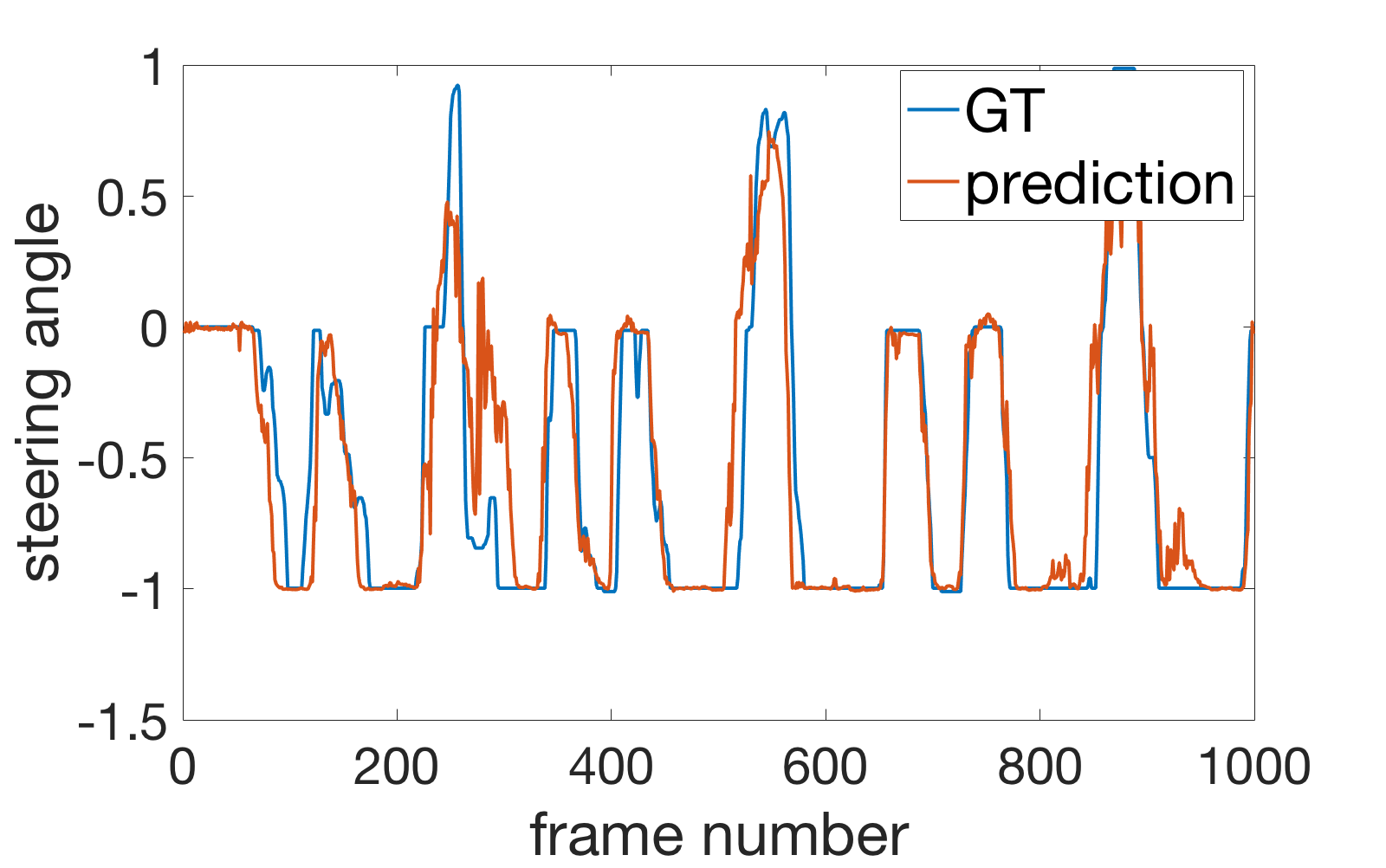}
			\caption{Proposed method}
		\end{subfigure}
	\end{center}
	\caption{Steering angle prediction of different method in one recordings of the indoor dataset.}
	\label{fig:indoor_steer}
\end{figure*}
In the indoor dataset, the remote control steering angle signal was normalised to $[-1, 1]$ ($-1$ corresponds to the leftmost and $1$ to the rightmost angle). 
\begin{figure}[h!]
	\centering
	\includegraphics[width=.24\textwidth]{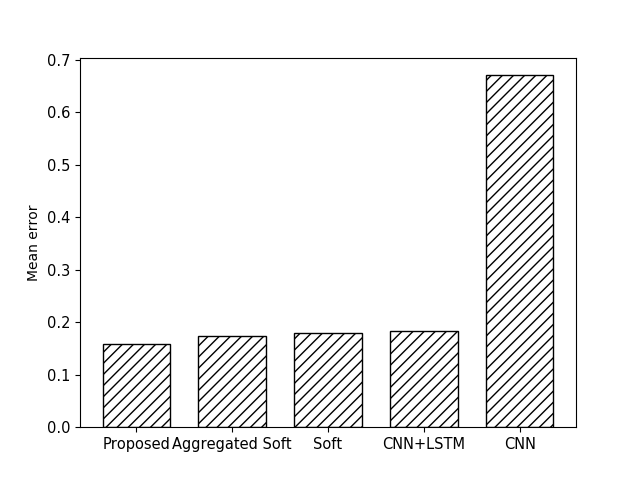}
	\includegraphics[width=.24\textwidth]{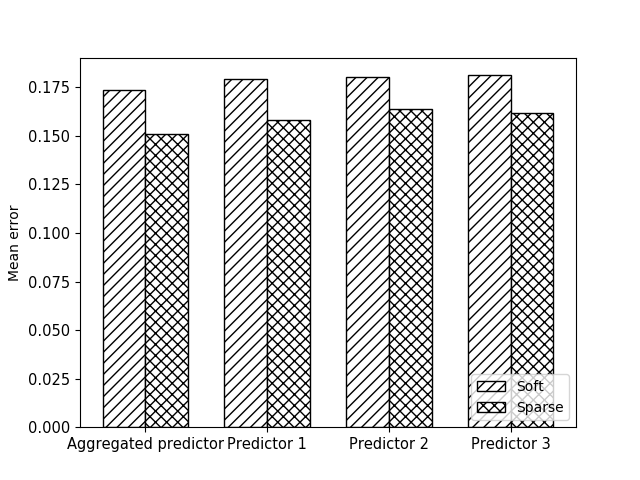}
	\caption{The mean error of different methods in the DIPLECS indoor dataset (left),and the mean error of each single predictor and aggregated predictor for sparse and soft attention in DIPLECS indoor dataset (right).}
	\label{fig:surrey_1}
\end{figure}
In Figures~\ref{fig:surrey_1}, we can see that the steering regression improves with the addition of any attention mechanism. Also, sparse attention performs better than soft attention, and the proposed aggregated sparse attention model performs best among those models. Importantly, we note that the models with attention provide a steering control that is not only more accurate but also smoother, which may be favourable for control applications. This particularly true for the proposed model. 
\begin{figure}[h!]
	\centering
	\includegraphics[width=.24\textwidth]{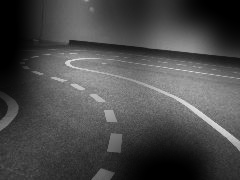}
	\includegraphics[width=.24\textwidth]{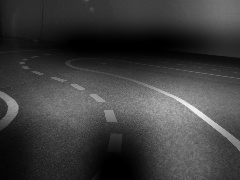}
	\caption{The attention map for soft attention (left) and sparse attention (right) in DIPLECTS indoor dataset}
	\label{fig:surrey_2}
\end{figure}
Figure~\ref{fig:surrey_2} show the attention maps for soft and sparse attention respectively. Note that the attended regions for most models appear to be focused on the road markings, some of them on the boundary marking or on the central one and some of them on both. We note also that the attention map for sparse attention is sparser than soft attention, which was the purpose of using the $\mathrm{sparsemax}$ function.

\subsection{Outdoor Dataset Results}
\begin{figure*}[h!]
	\begin{center}
		\begin{subfigure}{0.19\textwidth}
			\includegraphics[width=\textwidth]{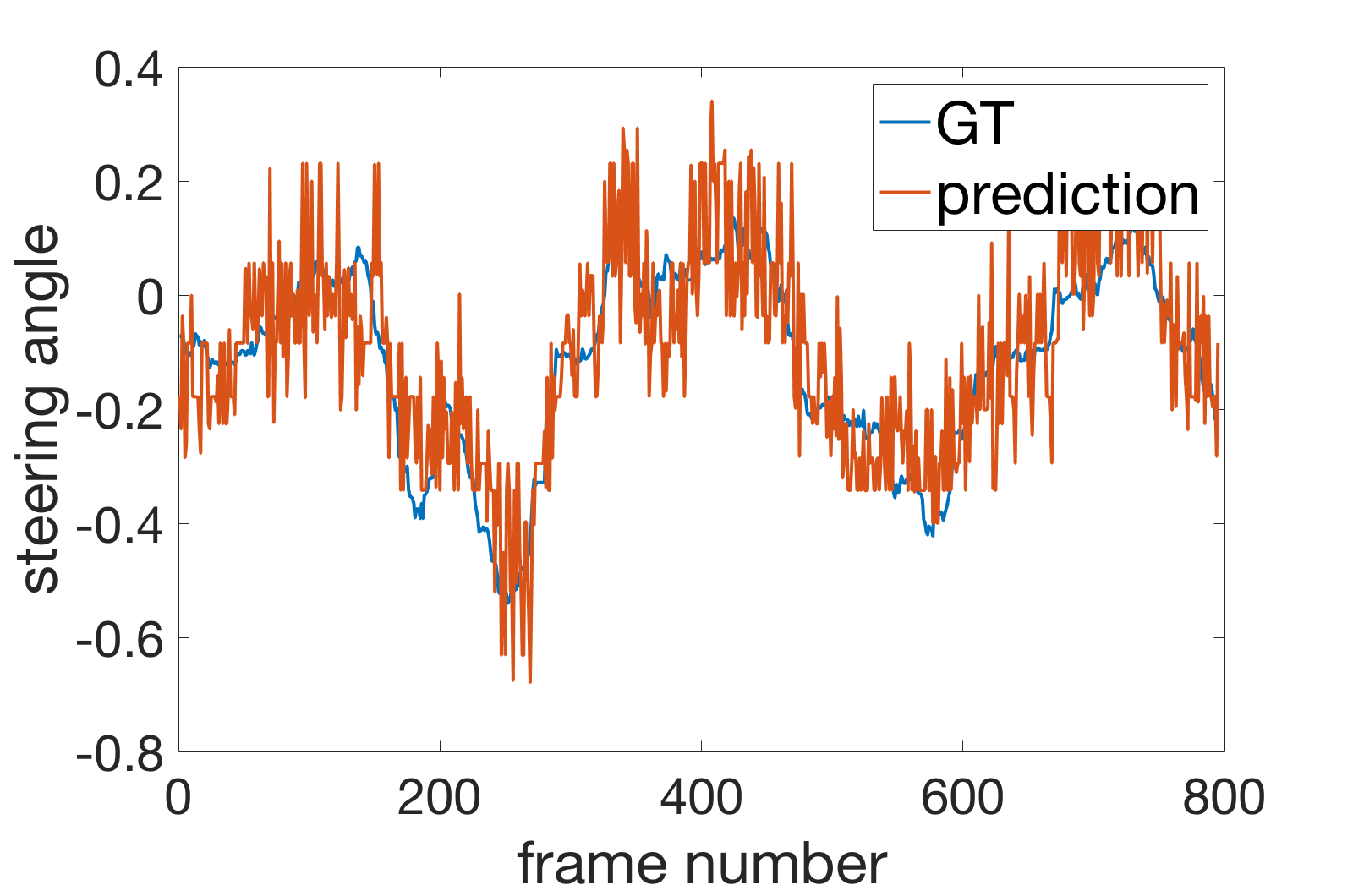}
			\caption{CNN+LSTM}
		\end{subfigure}
		\begin{subfigure}{0.19\textwidth}
			\includegraphics[width=\textwidth]{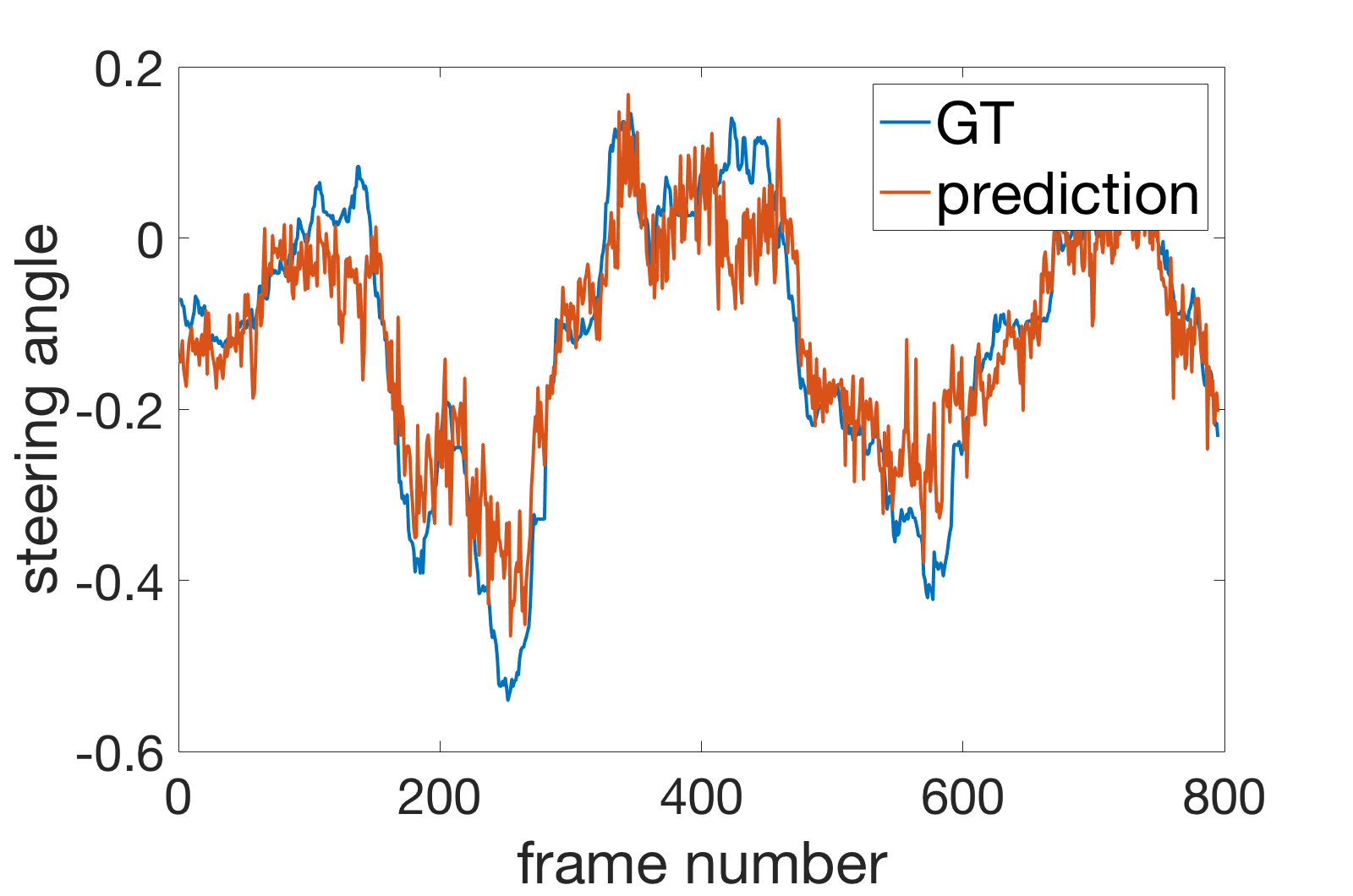}
			\caption{Soft attention}
		\end{subfigure}
		\begin{subfigure}{0.19\textwidth}
			\includegraphics[width=\textwidth]{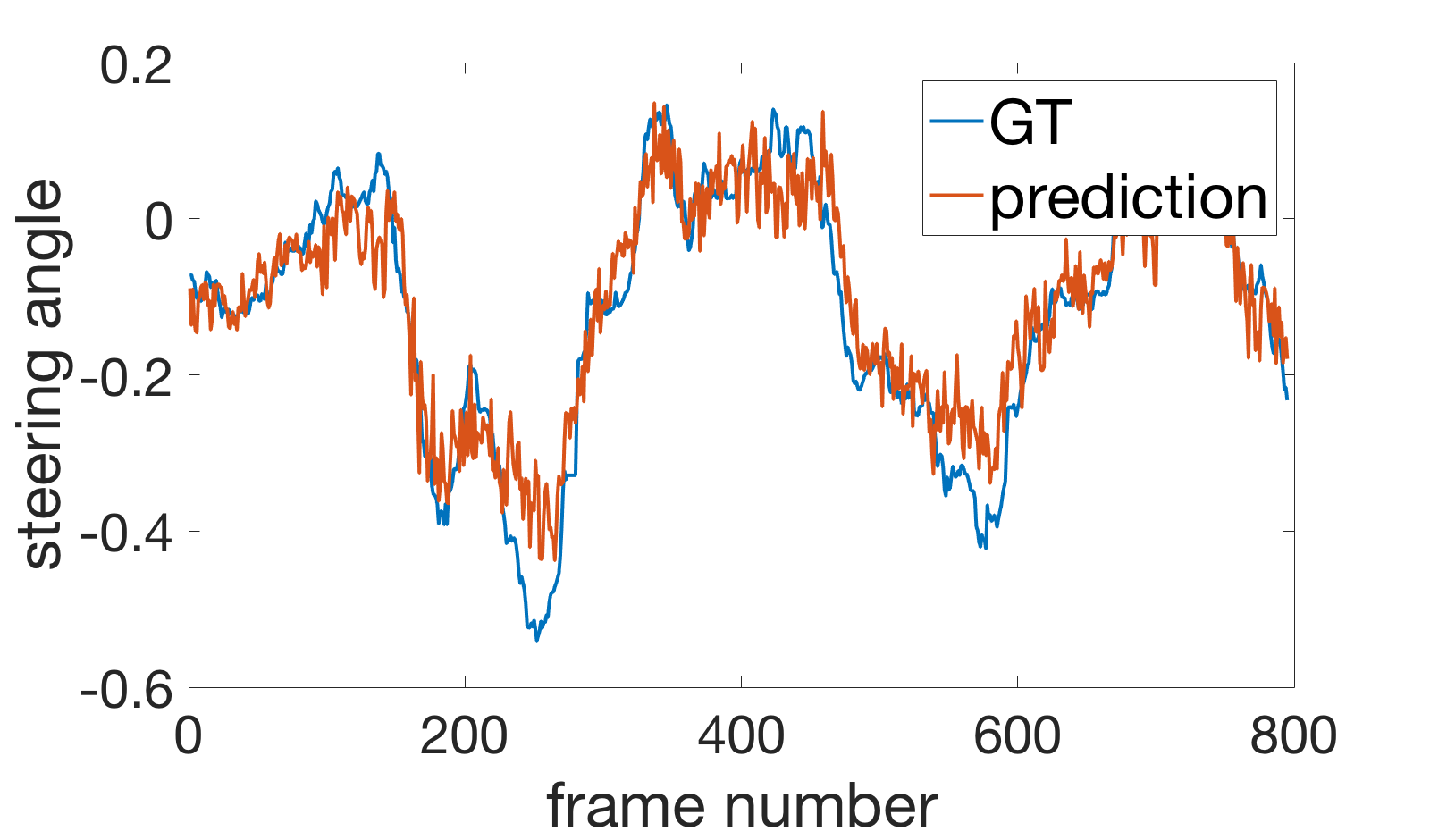}
			\caption{ASA}
		\end{subfigure}
		\begin{subfigure}{0.19\textwidth}
			\includegraphics[width=\textwidth]{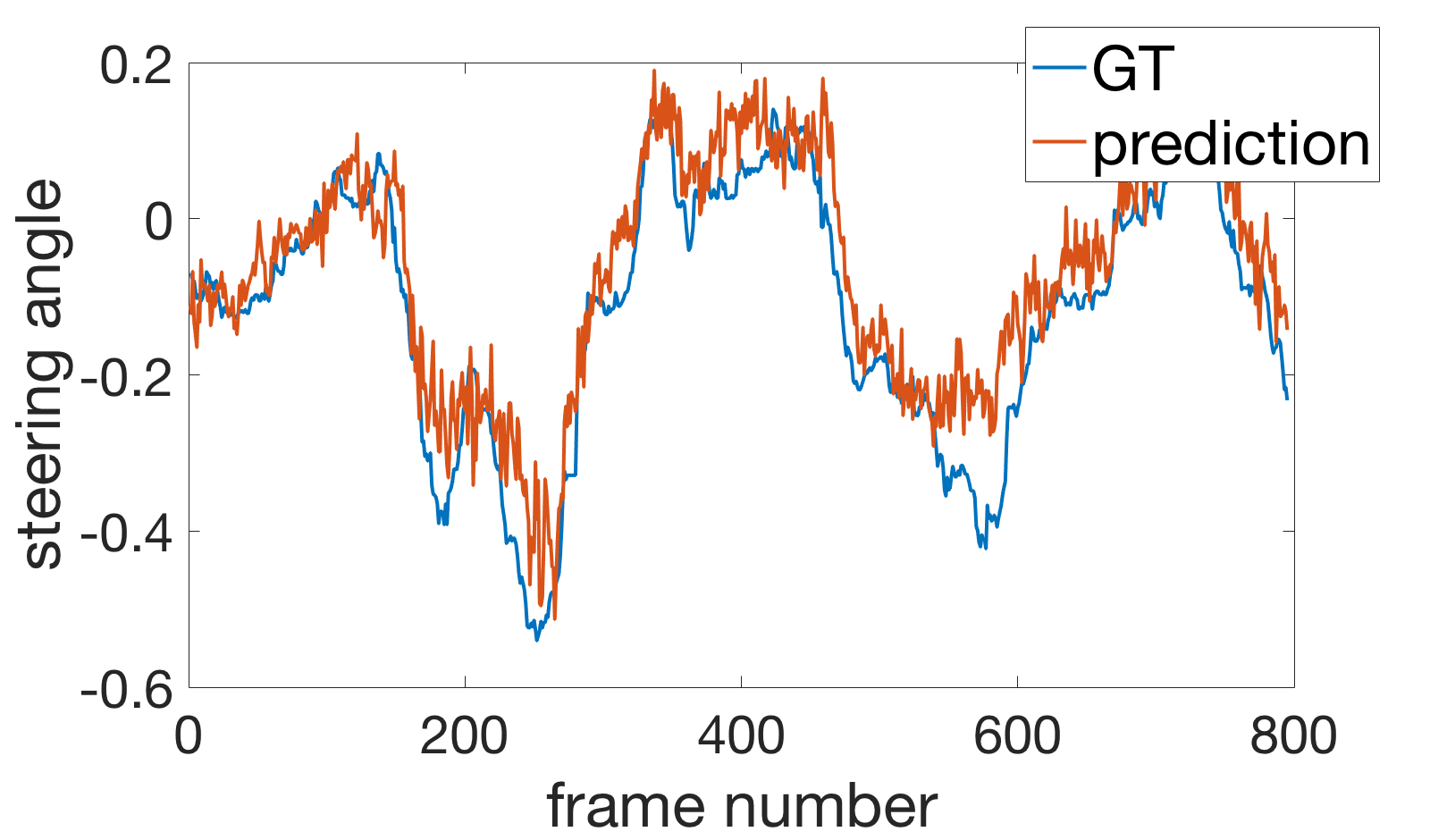}
			\caption{ASAR}
		\end{subfigure}
		\begin{subfigure}{0.19\textwidth}
			\includegraphics[width=\textwidth]{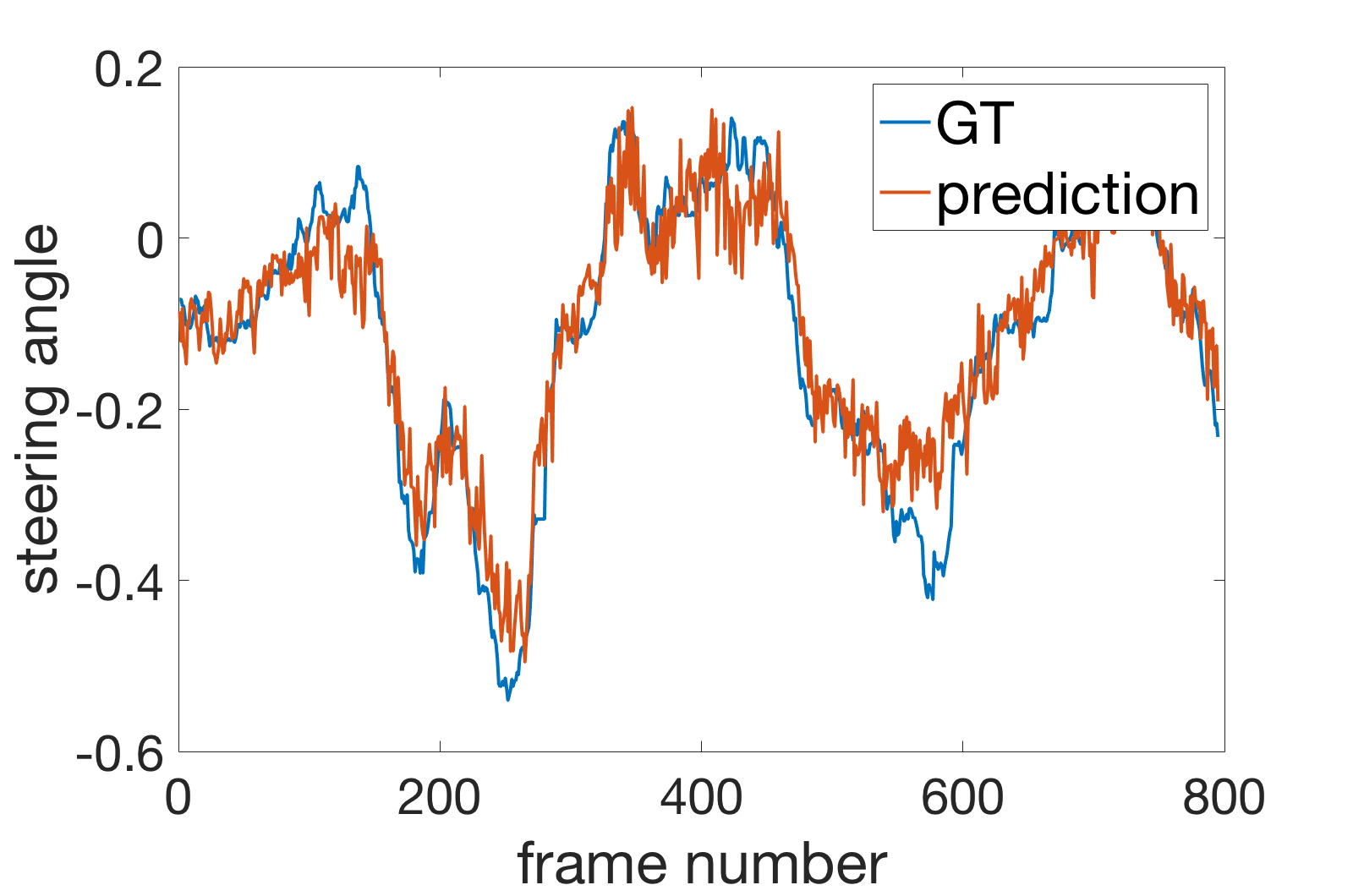}
			\caption{Proposed method}
		\end{subfigure}
	\end{center}
	\caption{Predicting the steering angle $0.5$ seconds later, by different methods for a subsequence of the DIPLECS outdoor testing dataset, the blue curve is the ground truth and the red one is the predicted steering angle.}
	\label{fig:surrey_steer}
\end{figure*}
In practice, a driver's actions are not instantaneous: due to reaction time, the driver's actions at any instant $t$ are based on the visual input received some time before. According to the studies in~\cite{green2000long, summala2000brake}, a driver's reaction time can varies from a few hundred milliseconds to several seconds. Before this section, we were predicting just the steering angle for the current frame ($s(t) = f(i(t))$). Here we use the current frame to predict the steering angle for different time delays ($s(t+d) = f(i(t))$.  We choose the delays corresponding to $0$s, $0.25$s, $0.5$s, $0.75$s, $1$s, and compare the results for different attention models.

\begin{figure}[h!]
	\centering
	\includegraphics[width=.35\textwidth]{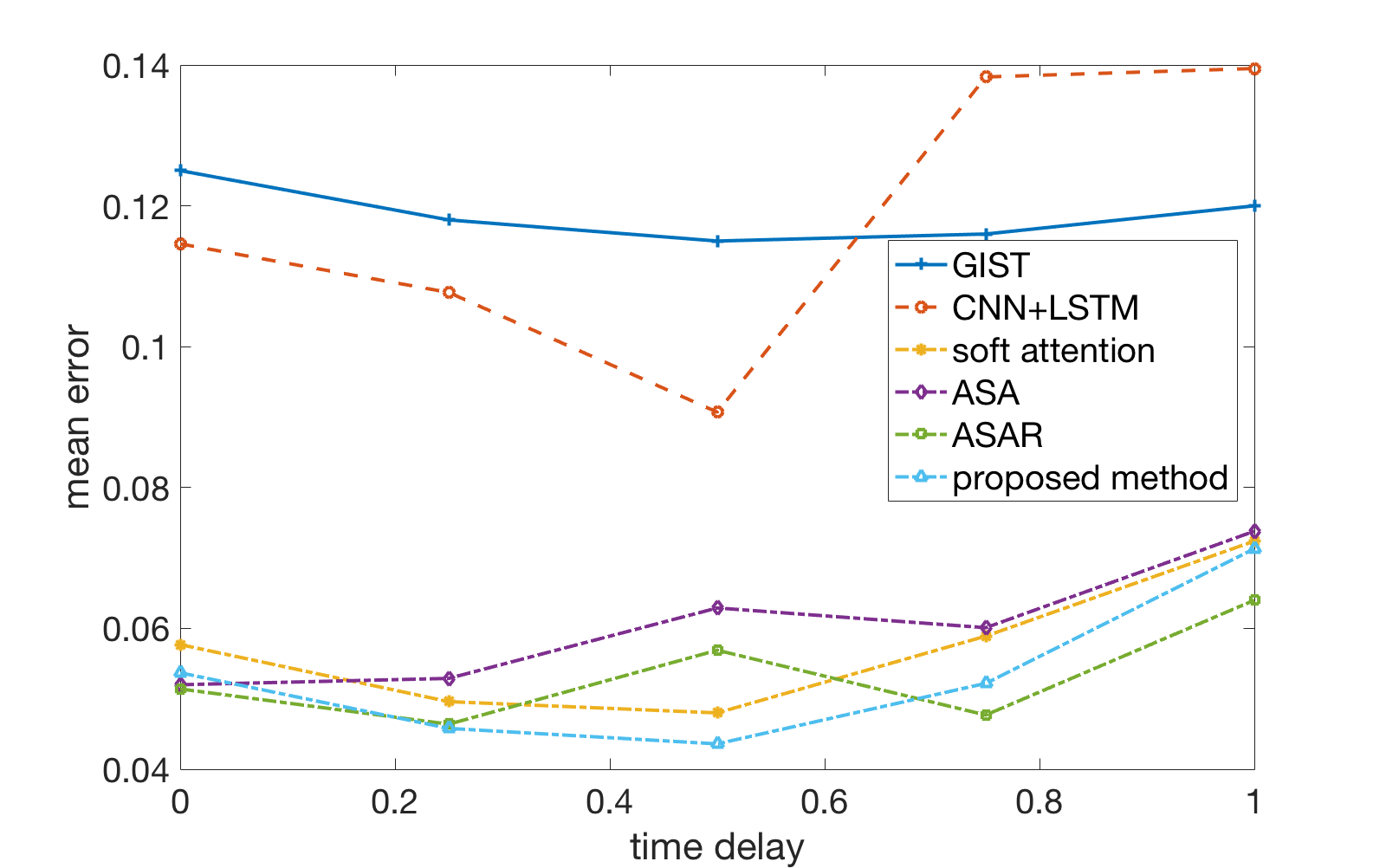}
	\caption{The mean regression error of different attention models on the DIPLECS outdoor testing dataset, for different time delay.}
	\label{fig:error_surrey}
\end{figure}
\begin{figure}[h!]
	\begin{tabular}{llll}
		Delay  & Attention map1                                                   & Attention map2 & Attention map3\\
		$0.25$ s  & \includegraphics[width=0.25\columnwidth]{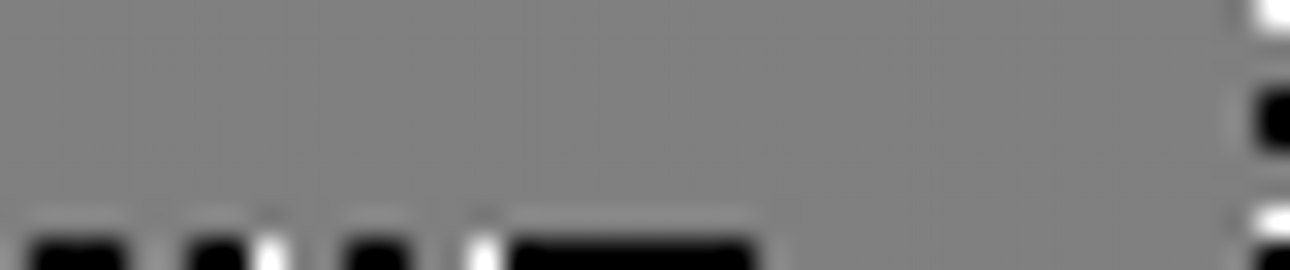} & \includegraphics[width=0.25\columnwidth]{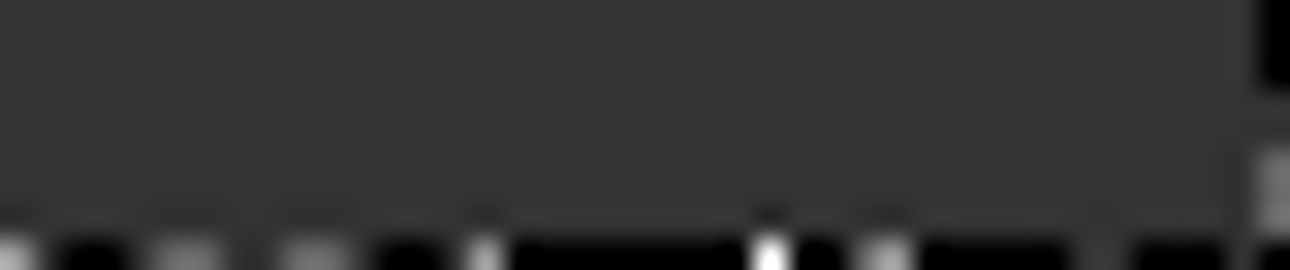} & \includegraphics[width=0.25\columnwidth]{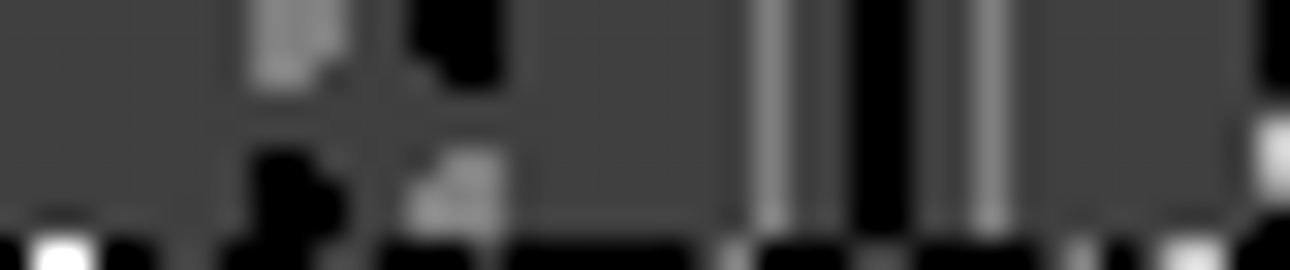} \\
		$0.5$ s  & \includegraphics[width=0.25\columnwidth]{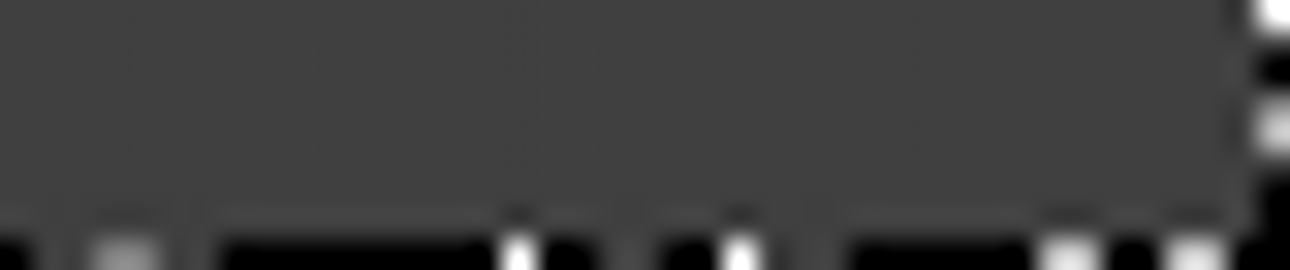} & \includegraphics[width=0.25\columnwidth]{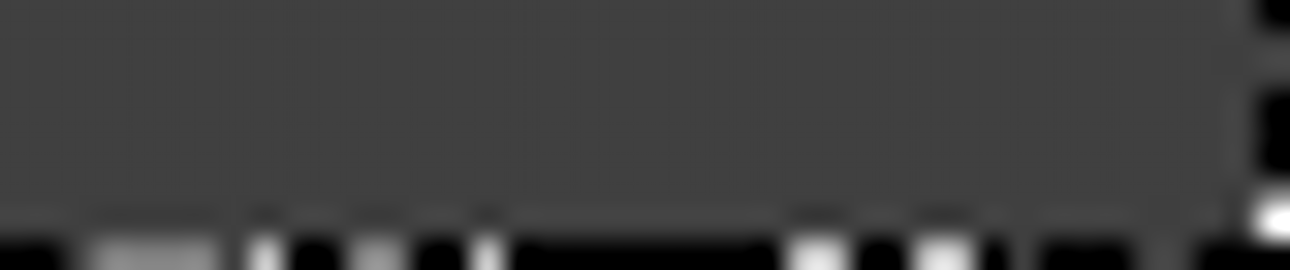} & \includegraphics[width=0.25\columnwidth]{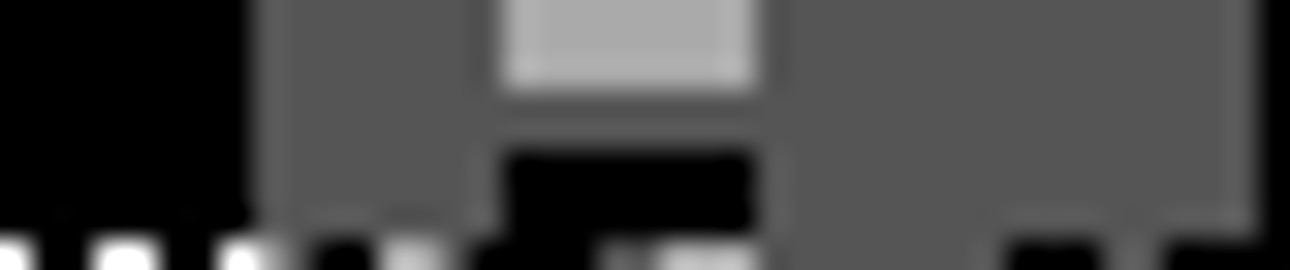} \\
	\end{tabular}
	\caption{Attention maps for $0.25s$ and $0.5s$ delays of each single sparse predictor}
	\label{fig:atten_10}
\end{figure}
\begin{figure}
	\centering
	\begin{subfigure}{0.155\textwidth}
		\centering
		\includegraphics[width=\textwidth]{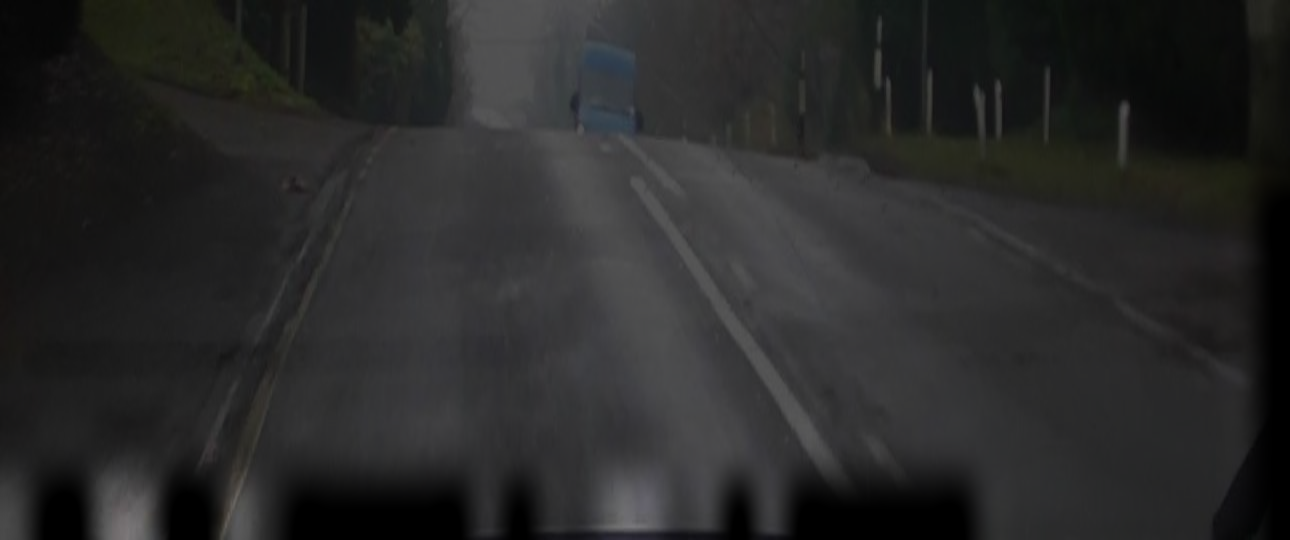}
		\caption{0s delay}
	\end{subfigure}
	\hfill
	\centering
	\begin{subfigure}{0.155\textwidth}
		\centering
		\includegraphics[width=\textwidth]{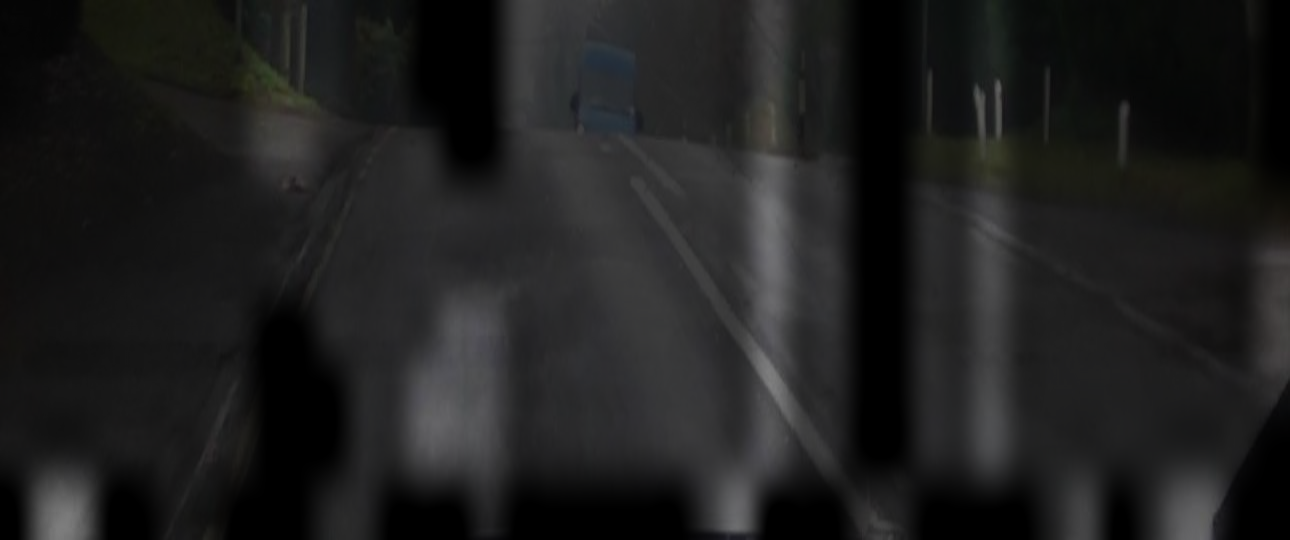}
		\caption{0.25s delay}
	\end{subfigure}
	\hfill
	\centering
	\begin{subfigure}{0.155\textwidth}
		\centering
		\includegraphics[width=\textwidth]{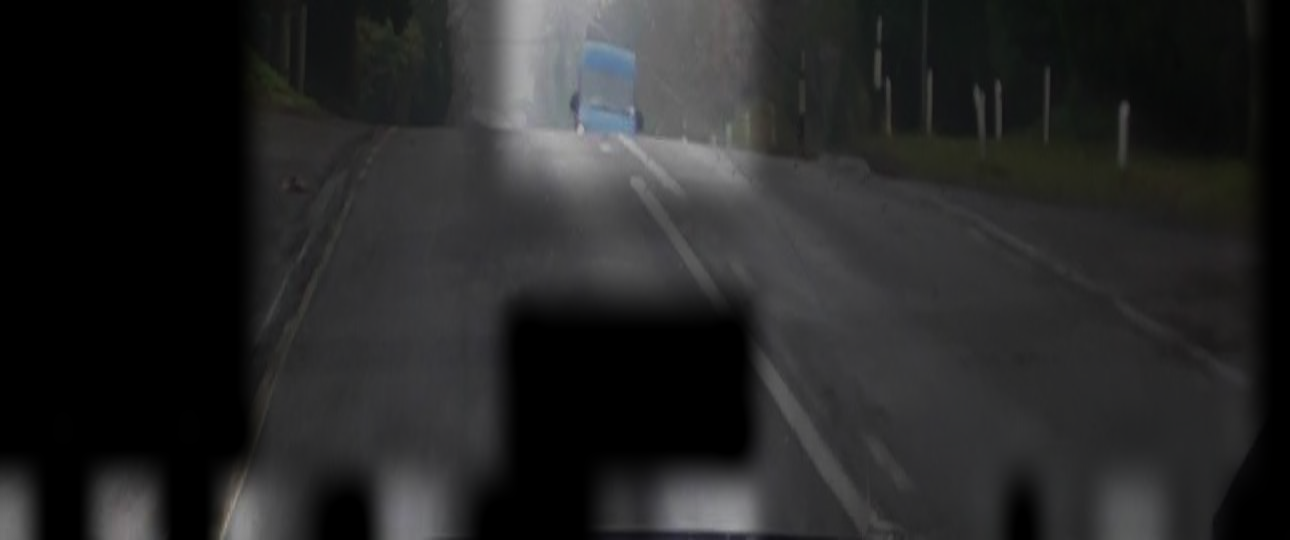}
		\caption{0.5s delay}
	\end{subfigure}
	\caption{The input frame overlapped by the attention map for different time delay.}
	\label{fig:forward}
\end{figure}

One can see from Figure~\ref{fig:error_surrey} that the proposed aggregated sparse attention model achieves the best performance among all methods with $0.5$s time delay, which is also the minimum mean error for all time delay among all methods. All models, which have an attention mechanism, perform much better than those without attention, and Figure~\ref{fig:surrey_steer} shows that the model with attention mechanism is more stable, with less perturbation for steering angle prediction, than the model without attention. Figures~\ref{fig:atten_10} and~\ref{fig:forward}, show the area of the visual field where attention is focused, for selected frames. Each single attention map only focus on a few different areas (the bright parts). Initially, attended locations are mostly at the bottom of the screen, but when time delay increases, we start seeing locations higher in the image being attended---this is especially visible for prediction time delay of $0.25s$ and $0.5s$.\\

\begin{figure}[h!]
	\centering
	\includegraphics[width=.24\textwidth]{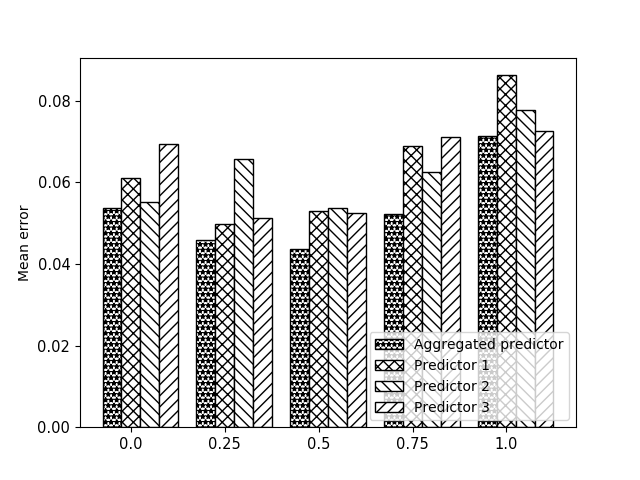}
	\includegraphics[width=.24\textwidth]{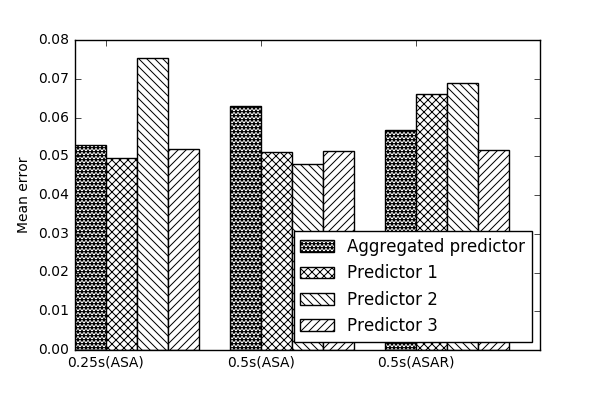}
	\caption{The mean error of each single predictor as well as aggregated predictor for different time delay of sparse (left) and soft(right) attention in DIPLECS outdoor dataset.}
	\label{fig:surrey_sparse}
\end{figure}

\begin{figure}[h!]
	\centering
	\begin{subfigure}{0.235\textwidth}
		\centering
		\includegraphics[width=\textwidth]{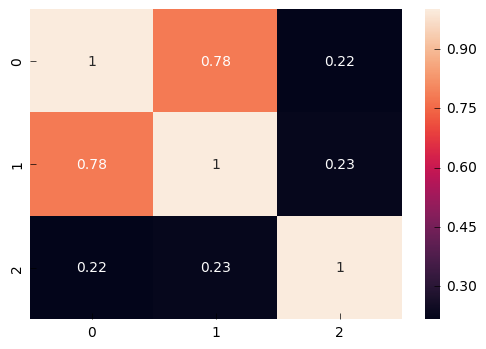}
		\caption{Soft attention}
	\end{subfigure}
	\hfill
	\centering
	\begin{subfigure}{0.235\textwidth}
		\centering
		\includegraphics[width=\textwidth]{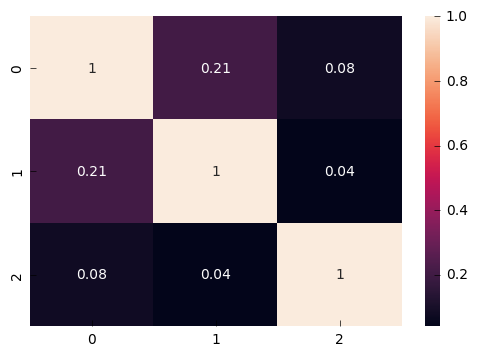}
		\caption{Sparse attention}
	\end{subfigure}
	\caption{The cross correlation between the attention map of each single attention model for aggregated sparse attention model and aggregated soft attention model with $0.5$s time delay.}
	\label{fig:correlation}
\end{figure}

It is also remarkable that in Figures~\ref{fig:surrey_sparse}, even though every single predictor within the aggregated sparse attention is trained using the same dataset, the aggregated model performs better than any single sparse attention predictor. For the aggregated soft attention model, two varieties of the model were compared: each single model within the aggregation has been trained on the same training set (ASA) or on different random subsets (ASAR). After model aggregation, the aggregated soft attention model performed worse than the single soft attention model for time delays $0.25$~s,$0.5$~s.
We suggest this is due to the cross-correlation between the attention maps of each single soft attention predictor. In Figure~\ref{fig:correlation}, one can see that there is a high correlation between the attention maps; even the smallest correlation coefficient of the soft attention map pairs is larger than the largest correlation coefficient for the sparse attention map, which suggests those soft attention model tend to focus on more or less the same region. In this case, if a single predictor model over- or underestimates the steering angle at some time, the other correlated predictors would also have the same trend in steering angle estimation, and after model aggregation it would negatively impact the final error. This result also confirms our suggestion about diversity of individual sparse attention maps, made in section \ref{Model Aggregation}.

\subsection{\textit{\textbf{Comma.ai}} Dataset}
\begin{figure*}[h!]
	\begin{center}
		\begin{subfigure}{0.19\textwidth}
			\includegraphics[width=\textwidth]{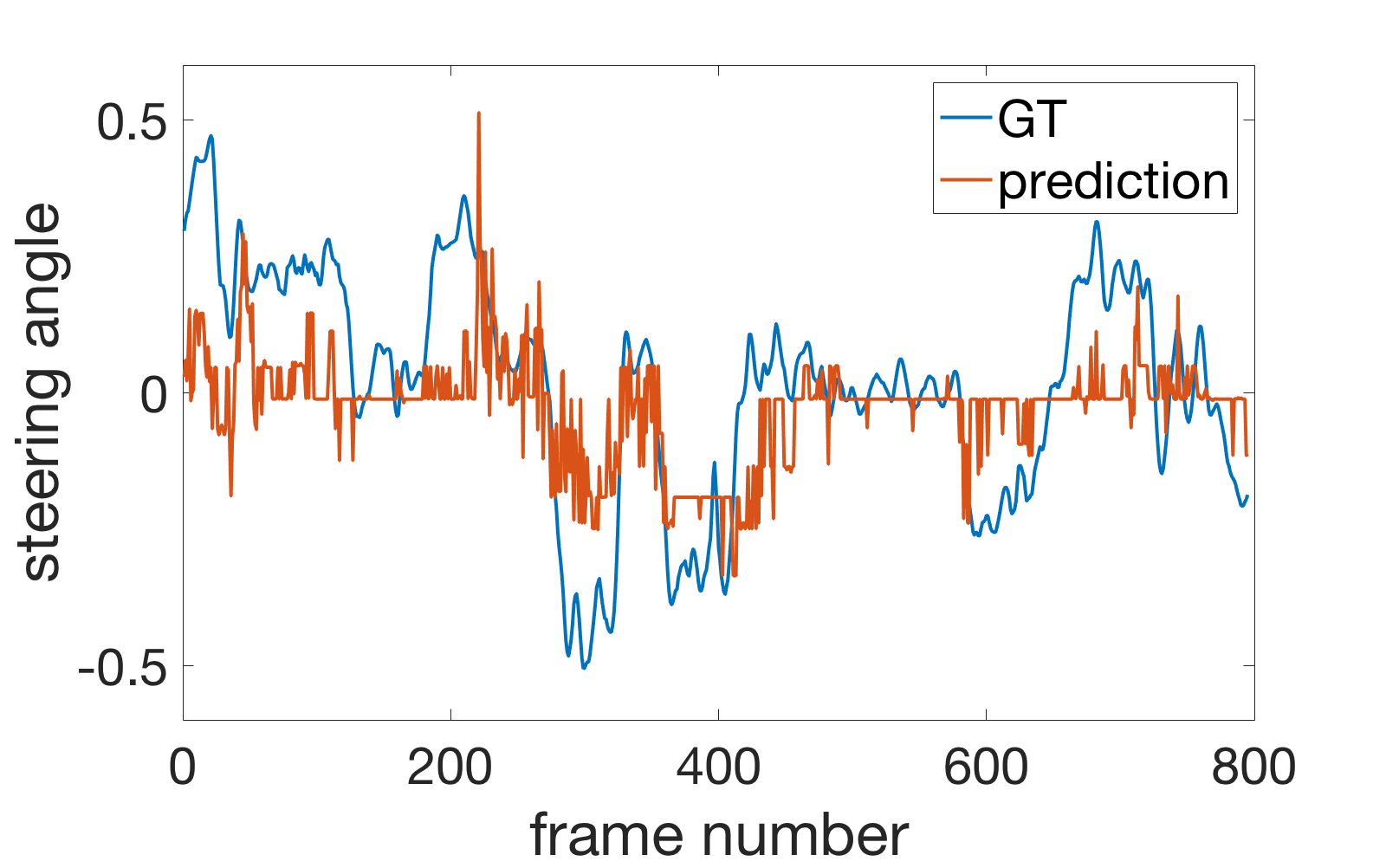}
			\caption{CNN+LSTM}
		\end{subfigure}
		\begin{subfigure}{0.19\textwidth}
			\includegraphics[width=\textwidth]{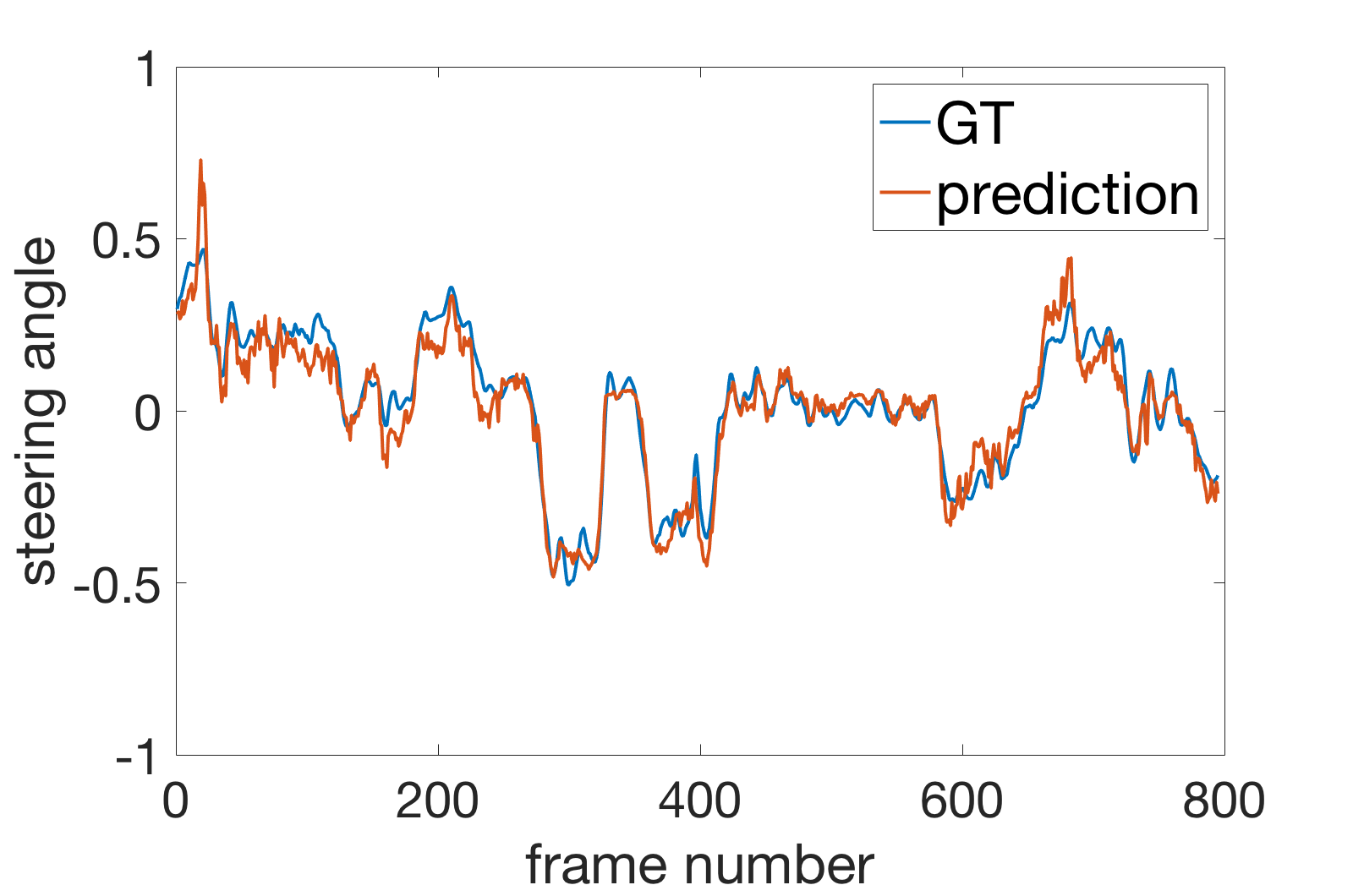}
			\caption{Soft attention}
		\end{subfigure}
		\begin{subfigure}{0.19\textwidth}
			\includegraphics[width=\textwidth]{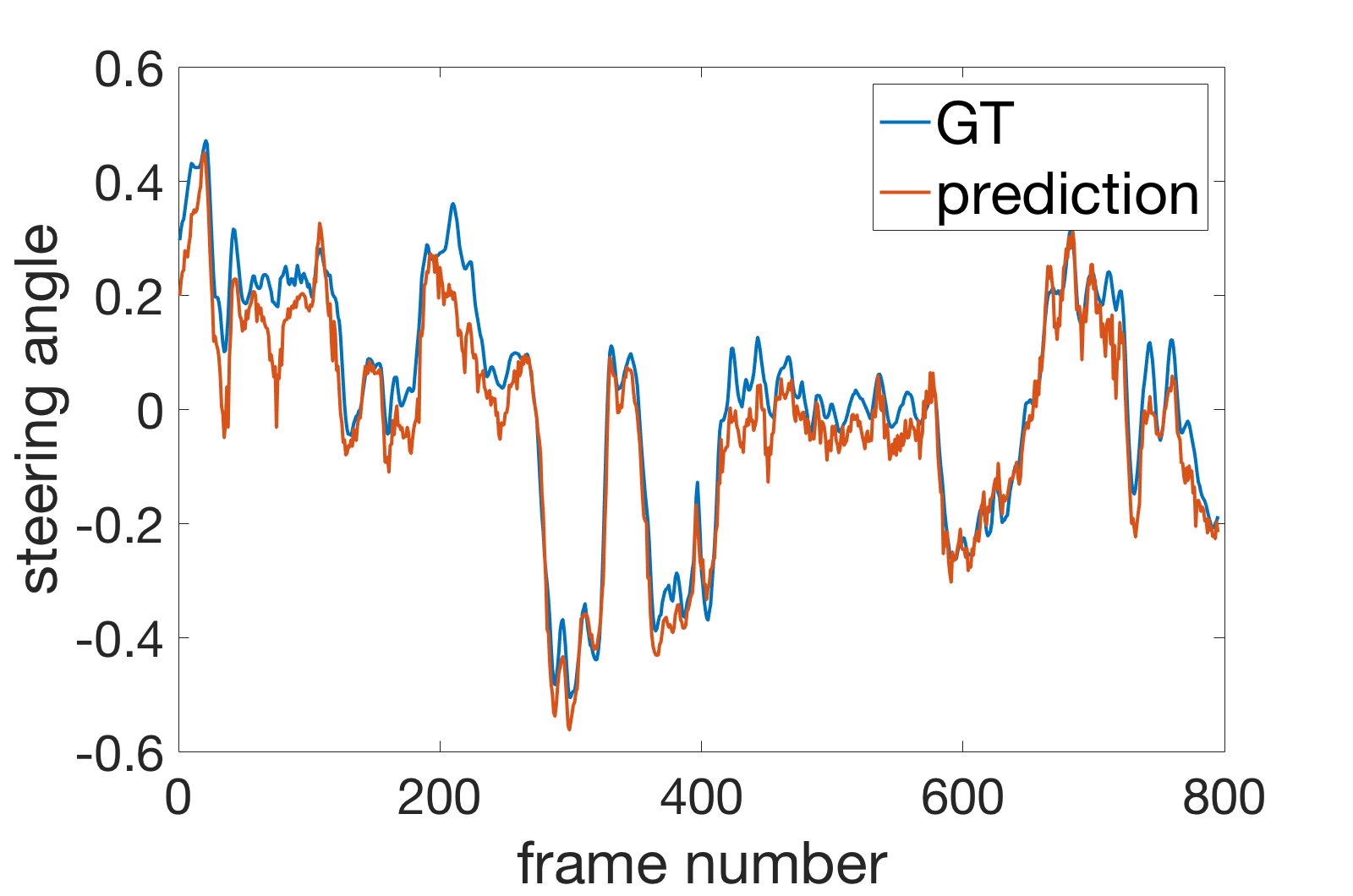}
			\caption{ASA}
		\end{subfigure}
		\begin{subfigure}{0.19\textwidth}
			\includegraphics[width=\textwidth]{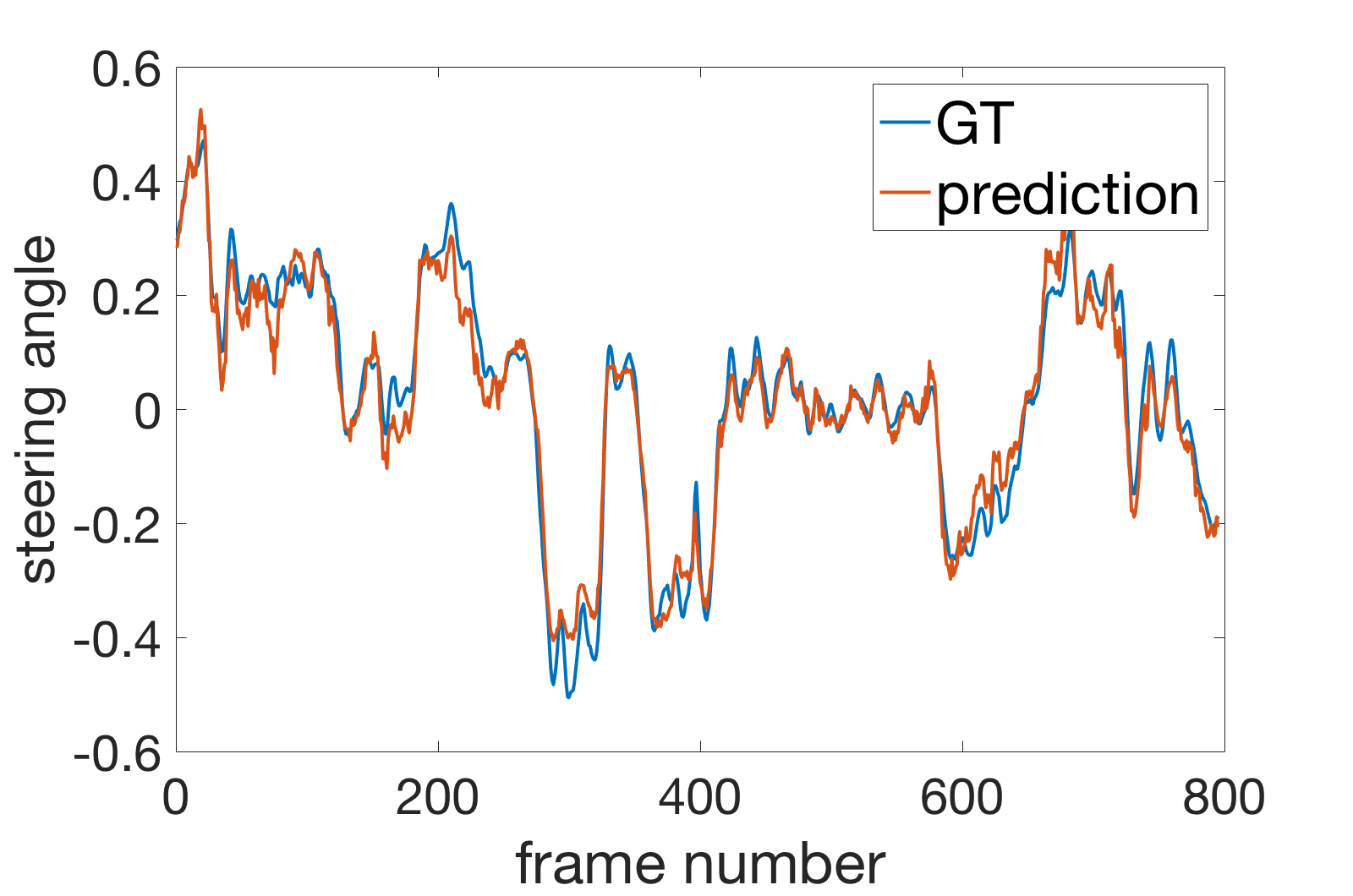}
			\caption{ASAR}
		\end{subfigure}
		\begin{subfigure}{0.19\textwidth}
			\includegraphics[width=\textwidth]{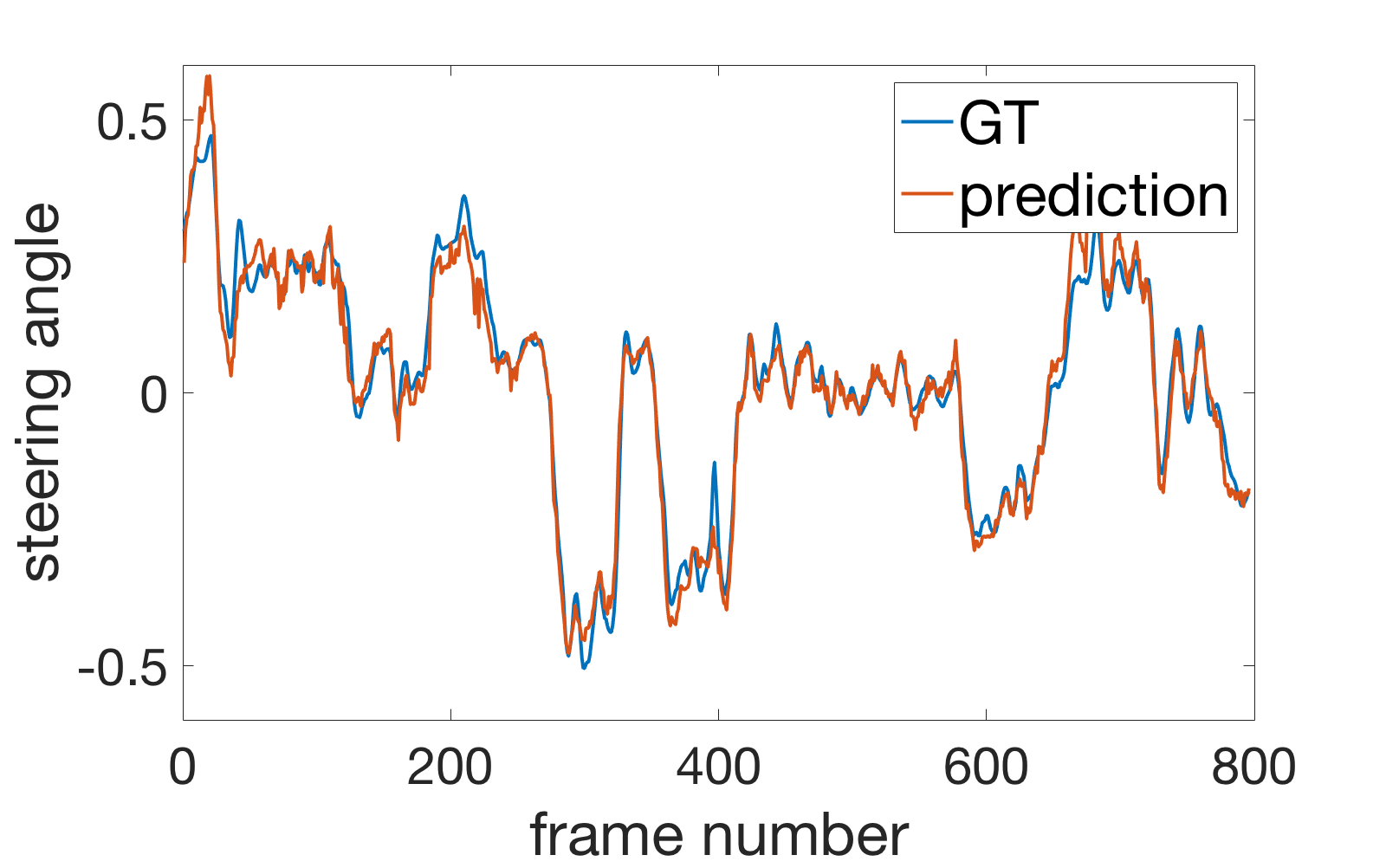}
			\caption{Proposed method}
		\end{subfigure}
	\end{center}
	\caption{Predicting the steering angle $1$ seconds later, by different methods for a subsequence of the \textit{Comma.ai} testing dataset, the blue curve is the ground truth and the red one is the predicted steering angle.}
	\label{fig:comma_steer}
\end{figure*}
\begin{figure}
	\centering
	\includegraphics[width=.35\textwidth]{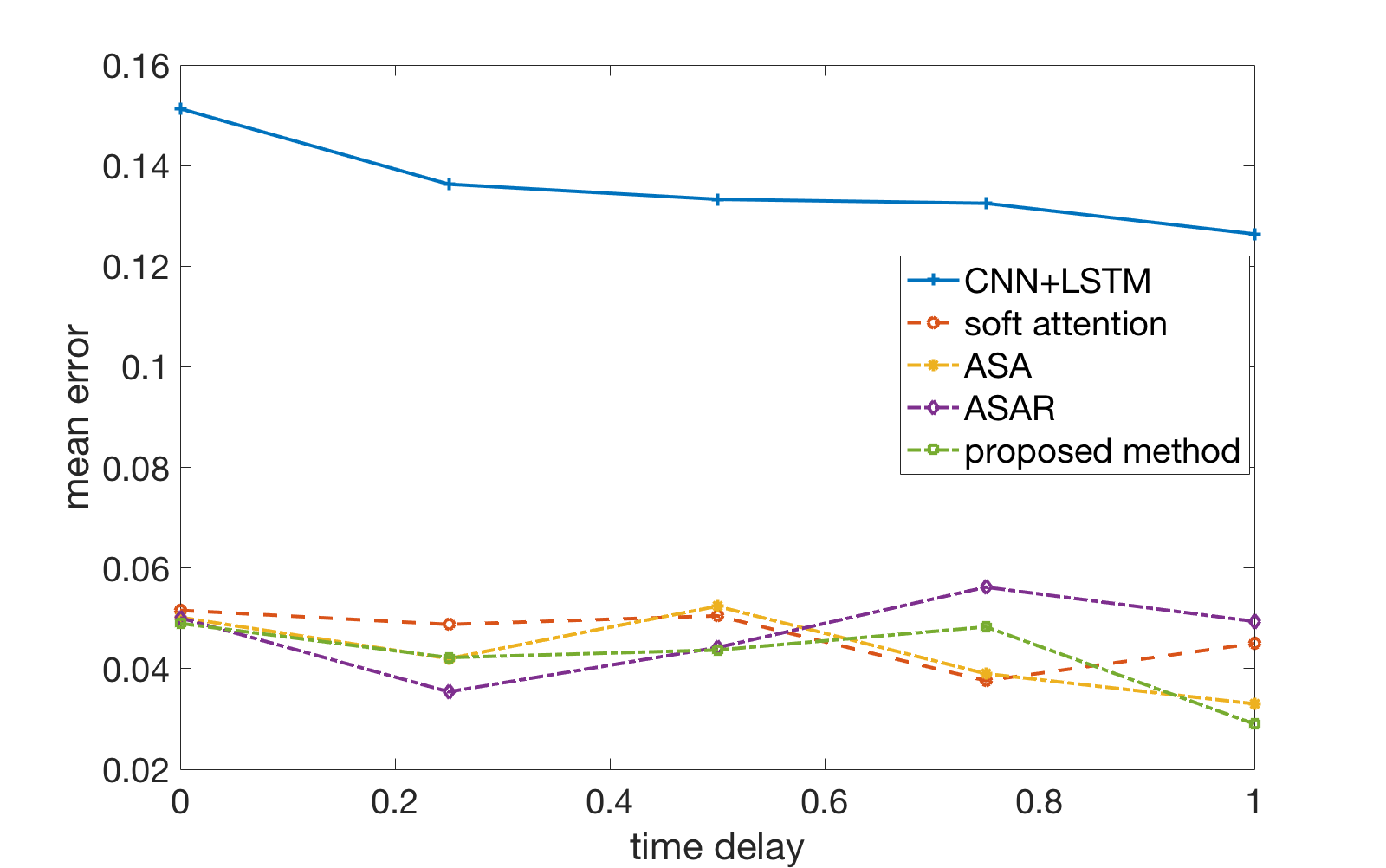}
	\caption{The mean error of different methods for different time delay  in \textit{Comma.ai} dataset.}
	\label{fig:comma_error}
\end{figure}
\begin{figure}
	\centering
	\includegraphics[width=.24\textwidth]{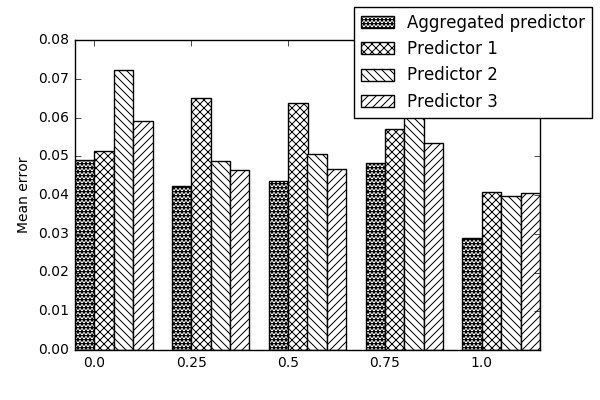}
	\includegraphics[width=.24\textwidth]{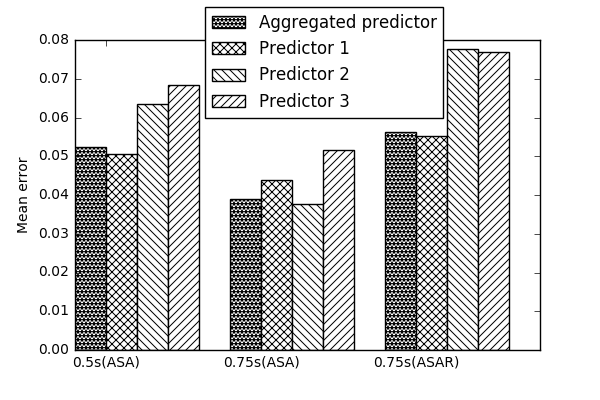}
	\caption{The mean error of each single predictor as well as aggregated predictor for different time delay of sparse (left) and soft (right) attention in \textit{Comma.ai} dataset.}
	\label{fig:si_sparse}
	\label{fig:si_soft}
\end{figure}

The testing procedure for \textit{Comma.ai} dataset is the same as for DIPLECS outdoor dataset. One can see from Figure~\ref{fig:comma_steer},~\ref{fig:comma_error} that the proposed aggregated sparse attention model still achieves the best performance among all methods, but for different time delay ($1$s time delay), we suggest that this is due to the driving environment being a highway with a broad view far ahead of the car, and therefore possibly requiring less attention from the driver than the countryside road in the DIPLECS outdoor dataset. On this dataset as previously, all models with attention mechanism perform much better than models without attention. 
Also, Figure~\ref{fig:si_sparse} (left) confirms, as was the case with the DIPLECS outdoor dataset, that even though every single predictor within the aggregated sparse attention is trained using the same dataset, the aggregated model performs significantly better than any single sparse attention predictor. The performance improvement from attention model aggregation is less evident when considering soft attention, as shown in Figure~\ref{fig:si_sparse} (right). 

\section{Conclusion}\label{sec:conclusion}
Attention plays an essential role in human driving. This article experiments with existing neural network models for task-directed attention, and proposed improved models the task of steering a car autonomously. 
Our experiments show that: i) all attention models improve steering prediction significantly; ii) a sparse attention model yields better performance than classical soft attention; and iii) an aggregated ensemble based on randomised attention models can achieve significantly better performances than a single attention model, even when trained on the same data. 
The method has been assessed in a variety of scenarios on three datasets, and achieves better performance than state-of-the-art. Additionally, as was done in previous published works the problem of steering angle prediction with a perception-action delay has been considered, demonstrating that the model achieves the best performance for $0.5$s delay for countryside road and $1$s delay for a highway.

\bibliographystyle{ieeetr}
\bibliography{bib.bib}

\end{document}